\documentclass{article}

  \usepackage[final]{neurips_2022}

\usepackage[utf8]{inputenc} 
\usepackage[T1]{fontenc}  
\usepackage{hyperref}    
\usepackage{url}      
\usepackage{booktabs}    
\usepackage{amsfonts}    
\usepackage{nicefrac}    
\usepackage{microtype}   
\usepackage{xcolor}     
\usepackage{lipsum}
\usepackage{amsmath}
\usepackage{graphicx}
\usepackage{subfigure}
\usepackage[utf8]{inputenc}

\newcommand{\AF}{\mathcal{A}}
\definecolor{customblue}{RGB}{79, 173, 235}

\title{Stochastic Adaptive Activation Function}

\author{%
Kyungsu Lee \\
DGIST\thanks{Daegu Gyeongbuk Institute of Science and Technology}\\
42988 Daegu, South Korea \\
\texttt{ks\_lee@dgist.ac.kr} \\
\And
Jaeseung Yang \\
DGIST \\
42988 Daegu, South Korea \\
\texttt{yjs6813@dgist.ac.kr} \\
\AND
Haeyun Lee \\
DGIST, SAMSUNG SDI \\
17084 Yong-In, South Korea \\
\texttt{haeyun.lee@samsung.com} \\
\And
Jae Youn Hwang\thanks{corresponding author} \\
DGIST \\
42988 Daegu, South Korea \\
\texttt{jyhwang@dgist.ac.kr}
}

\begin{document}

\maketitle

\setcounter{footnote}{0}

\begin{abstract}
The simulation of human neurons and neurotransmission mechanisms has been realized in deep neural networks based on the theoretical implementations of activation functions. However, recent studies have reported that the threshold potential of neurons exhibits different values according to the locations and types of individual neurons, and that the activation functions have limitations in terms of representing this variability. Therefore, this study proposes a simple yet effective activation function that facilitates different thresholds and adaptive activations according to the positions of units and the contexts of inputs. Furthermore, the proposed activation function mathematically exhibits a more generalized form of Swish activation function, and thus we denoted it as Adaptive SwisH (ASH). ASH highlights informative features that exhibit large values in the top percentiles in an input, whereas it rectifies low values. Most importantly, ASH exhibits trainable, adaptive, and context-aware properties compared to other activation functions. Furthermore, ASH represents general formula of the previously studied activation function and provides a reasonable mathematical background for the superior performance. To validate the effectiveness and robustness of ASH, we implemented ASH into many deep learning models for various tasks, including classification, detection, segmentation, and image generation. Experimental analysis demonstrates that our activation function can provide the benefits of more accurate prediction and earlier convergence in many deep learning applications.
\footnote{Our code is available at https://github.com/kyungsu-lee-ksl/ASH}
\end{abstract}

\section{Introduction}
Searching for the optimal activation functions has been a challenge in the field of artificial intelligence~\citep{maas2013rectifier, ramachandran2017searching, clevert2015fast}. Early activation functions have been studied to compensate for the non-linearity of the artificial neural networks or ameliorate the gradient vanishing problem~\citep{hertz1997nonlinear, hochreiter1998recurrent}. Recently, novel activation functions have been suggested with the zero-centered or parametric properties that improve the training efficiency of deep neural networks (DNNs)~\citep{maas2013rectifier, clevert2015fast}. Currently, the activation functions focusing on the stability of DNNs or probabilistic distribution of inputs have been proposed~\citep{hendrycks2016gaussian, misra2019mish}. Advances in the activation functions have allowed DNNs to perform various tasks such as detecting or segmenting target objects in sophisticated images or even generating new images beyond the simple classifiers~\citep{simonyan2014very, zhao2019object, badrinarayanan2017segnet, goodfellow2014generative}.

The activation functions has evolved to behave more like a human neuron~\citep{sharma2017activation,lee2017deep}. However, \citet{izhikevich2003simple, evans2018synaptic} reported that the neurotransmission mechanism, including the membrane, action, and threshold potentials of human neurons, is subject to the location or the connection type of the neurons. Additionally, humans perceive objects with surrounding contexts using the $N:N$ mapping of visions to human neurons rather than the 1:1 mapping of a pixel to an input node in a neural network~\citep{liu2018deep}. The connections between neurons can be realized through the linear combinations of layers in DNNs. However, DNNs have limitations in terms of realizing contextual perception. This implies that the further improvement in deep neural networks and convolutional neural networks (CNNs) can be realized despite the impressive performance on image analysis~\citep{jinsakul2019enhancement, misra2019mish}. Therefore, the development of DNNs is leaned to mimic human perception by realizing the mechanism of human neurons~\citep{aggarwal2018neural, lindsay2021convolutional}.

Currently, the primary issue is that many activation functions exhibit passivity, in this paper, indicating that they determine outputs only concerning the value of one element rather than entire contexts. For instance, the Rectified Linear Unit (ReLU), defined as $f(x)=\text{max}(x, 0)$, determines the output values related to $x$~\citep{fukushima1982neocognitron}, whereas the \textit{softmax} function generates output values as the ratio of the input value to the totals~\citep{goodfellow2017deep}. Particularly, ReLU exhibits passivity, whereas \textit{softmax} does not. Suppose an image can be classified by considering 80\% of the total portion. Current activation functions are limited in terms of classifying such an image since only elements (pixels) of the image are used to rectify the image rather than a ratio. Another limitation is that the activation functions are invariant. Although the parametric activation functions update their parameters during training, the resulting parameters are invariant during the inference phase~\citep{xu2015empirical, bingham2022discovering}. Therefore, the limited rectification can be realized by the invariant parameters or thresholds regardless of new inputs from different domains (e.g., test set).

\paragraph{Contribution}
To realize the mechanism of human neurons that rectify inputs considering their contexts, we propose a novel ASH activation function. The main contributions of this study are to suggest a simple yet effective activation function, ASH, and to implement ASH in a mathematically effective form. Going beyond the passive activation functions, ASH activation function is designed as (1) an active activation function that provides outputs regarding the context of inputs and (2) a conditional activation function that employs an adaptive threshold. Unlike ReLU or Leaky ReLU, the threshold value of ASH is adaptively changed by considering the contextual information.
\begin{equation}
  f(x) = \begin{cases}
  x & \text{if}\ x \geq \theta, \\
  0 & \text{otherwise}
  \end{cases}
\end{equation}
In particular, the threshold value ($\theta$) is adaptively changed according to the input distribution without heavy calculation, and thus ASH provides outputs considering the contexts of inputs adaptively. By applying ASH, we obtained the following theoretical and experimental results:

\begin{itemize}
  \item We conducted mathematical modeling on ASH in an effective form to ensure trainable and parametric properties, and thus ASH exhibits parametric and adaptive properties. The baseline threshold of ASH is initially trained during the training phase, and the threshold value is adaptively fine-tuned according to the contexts of inputs without heavy calculations.
  \item We theoretically verified that ASH adaptively changes its threshold alongside the stochastic distribution of inputs. This implies that ASH provides outputs regarding the entire contexts of inputs, thus leading to enhanced feature extraction.
  \item We theoretically verified that ASH exhibits a generalized formula of Swish activation function and provided the mathematical explanations for the superior performance of Swish, which was empirically searched in the previous work.
  \item We experimentally showed that ASH improves the performance of deep learning models on various tasks and shortens the convergence epoch.
\end{itemize}

\paragraph{Related works}
Activation functions affect the performance of the training process to determine a functional subspace of a DNN~\citep{hayou2019impact}. In particular, the non-linearity using activation functions have been introduced to prevent the issue of the linear transformation causing simple feature extractions in the DNNs~\citep{misra2019mish,jarrett2009best}. DNNs with non-linearity have been employed to 
perform complex tasks ~\citep{leshno1993multilayer}. In the early era, Rectified Linear Unit (ReLU) replaced the classical activation functions such as sigmoid and tanh~\citep{nair2010rectified} due to its simple and computational efficiency compared to other activation functions. During decades, many activation functions have been proposed , including Leaky ReLU \citep{maas2013rectifier}, Exponential Linear Unit (ELU) \citep{clevert2015fast}, Gaussian Error Linear Unit (GELU) \citep{hendrycks2016gaussian}, Scaled Exponential Linear Unit (SELU) \citep{klambauer2017self}, and Swish \citep{ramachandran2017searching} to improve the performance and stability of learning parameters in DNNs. Those activation functions have solved the dying ReLU problem, which exhibits a zero value in the negative region, and improved DNNs more smoothly for stable optimization. In particular, ELU and SELU have realized internal normalization in the layer using the zero-mean property. GELU exploited a Gaussian error and could implement an adaptive dropout to apply a higher probabilistic intuition.

\paragraph{Problem statement}
For adaptive thresholding, ASH exploits a stochastic selection methodology such as a quick selection~\citep{hoare1961algorithm}. In particular, for enhanced feature extraction, the informative elements, which exhibit large values, should be identified as an attention mechanism. In contrast, some elements, which exhibit low relevance, should be required to be rectified. To this end, ASH is designed to identify informative elements but rectify others as 0.

Let $X \in R^{H \times W \times C}$ be a tensor (i.e., feature-map) disregarding the batch, but with a height ($H$), width ($W$), and channel ($C$), and let $\bar{X} \ni X$ be a set of feature-maps. Let $\AF$ be an activation function such that $\AF: \bar{X} \rightarrow \bar{X}$. We can then extract the $i^{th}$ element from $X$, and denote it as $x^{(i)}$. Here, the goal of this study is to design the activation function represented as follows:

\begin{equation}
\label{eq:def_af}
  \AF(x^{(i)}) =
  \begin{cases}
  x^{(i)} & \text{if}\ x^{(i)}\ \text{is ranked in the top-k percentile of X}, \\
  0 & \text{otherwise}
  \end{cases}
\end{equation}

Note that, the novel activation function $\AF$ is represented similar to ReLU, whereas its threshold is not invariant in contrast to ReLU, but it is subjected to the distribution of the input $X$. To simplify, let $C(x^{(i)}, k; X)$ be a condition whether $x^{(i)}$ is ranked in the top $k$ percentile of $X$, and the negation of $C$ is denoted as $\neg C$. In particular, the elements that satisfy $C(x^{(i)}, k; X)$ are from the first largest element to the ($0.01kN$)-th largest element in $X$, where $N$ indicates the number of elements in $X$. Here, the elements can be extracted using a simple algorithm as a quick selection. Suppose a set $\hat{X}$ that includes all elements in $X$ such that $\hat{X} = \{x^{(1)}, x^{(2)}, ..., x^{(N)}\}$. We can then construct subsets of $\hat{X}$ as $\hat{X}_C = \{x^{(i)} \in X | C(x^{(i)}, k; X)\}$ and $(\hat{X}_C)^c = \{x^{(i)} \in X | \neg C(x^{(i)}, k; X)\}$. Equation (\ref{eq:def_af}) can be then simplified as follows:

\begin{equation}
\label{eq:def_af_2}
  \AF(x^{(i)}) =
  \begin{cases}
  x^{(i)} & \text{if}\ x^{(i)} \in \hat{X}_C, \\
  0 & \text{otherwise}
  \end{cases}
\end{equation}

\begin{figure}[h]
\label{fig:activations}
\centering
\subfigure[Feature-map]{
\includegraphics[width=0.165\linewidth]{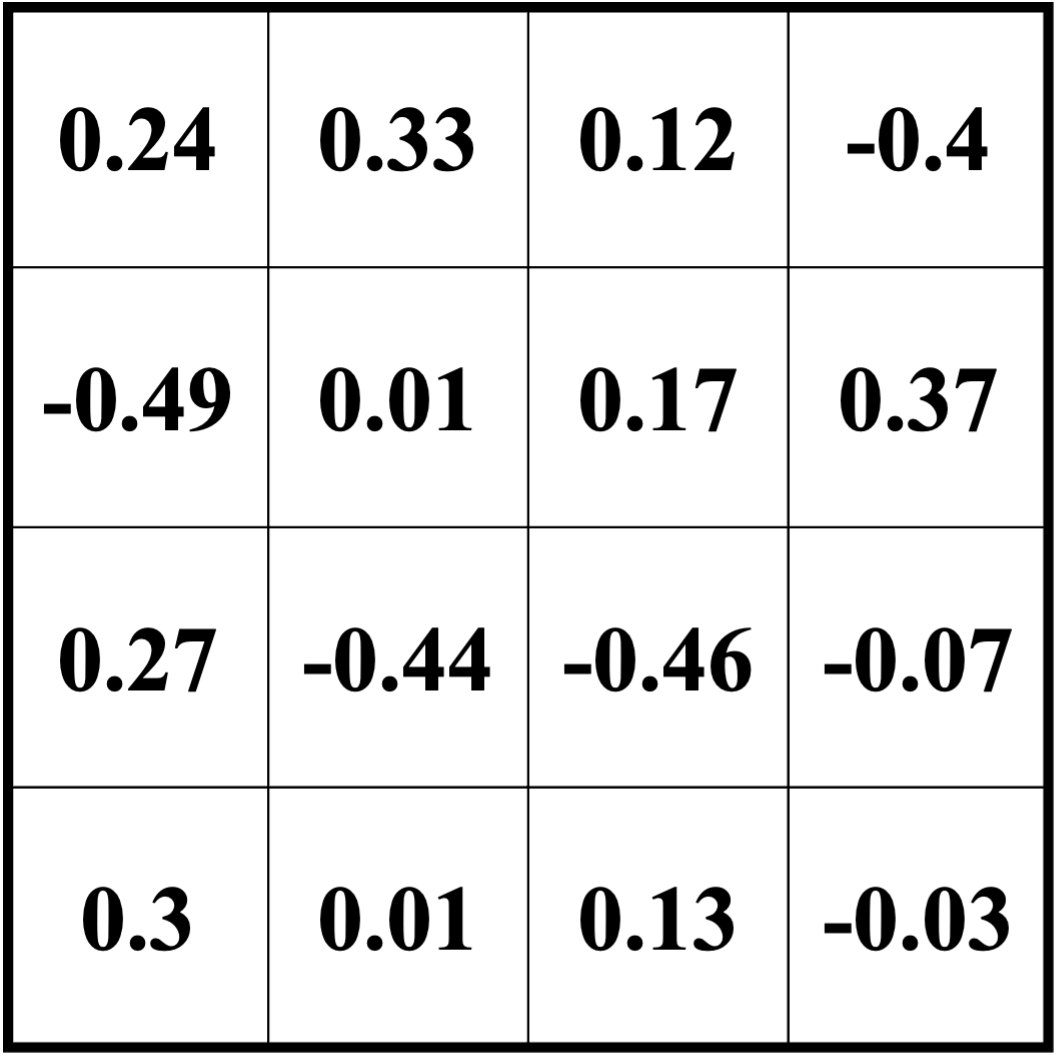}
\label{fig:activation_input}
}
\subfigure[ReLU]{
\includegraphics[width=0.165\linewidth]{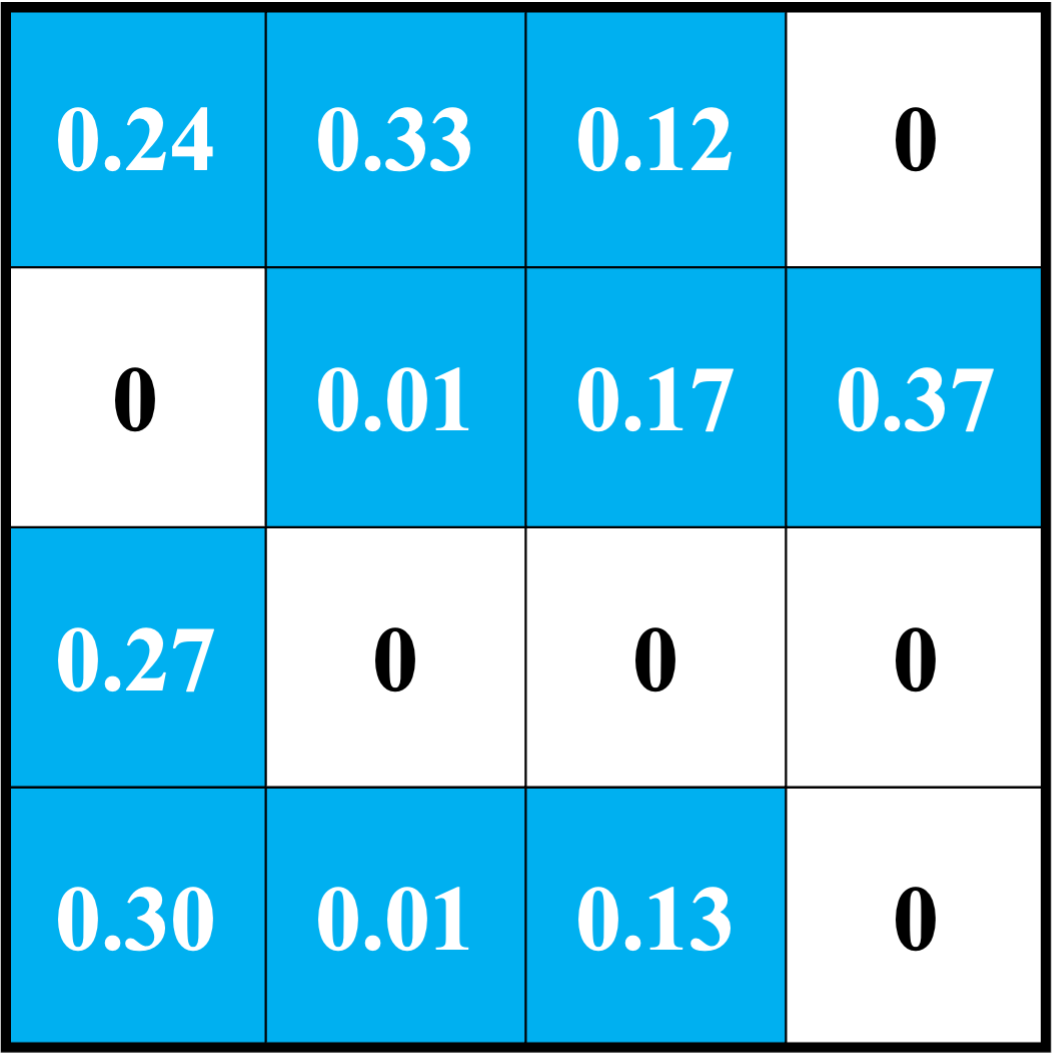}
\label{fig:activation_relu}
}
\subfigure[ASH (80\%)]{
\includegraphics[width=0.165\linewidth]{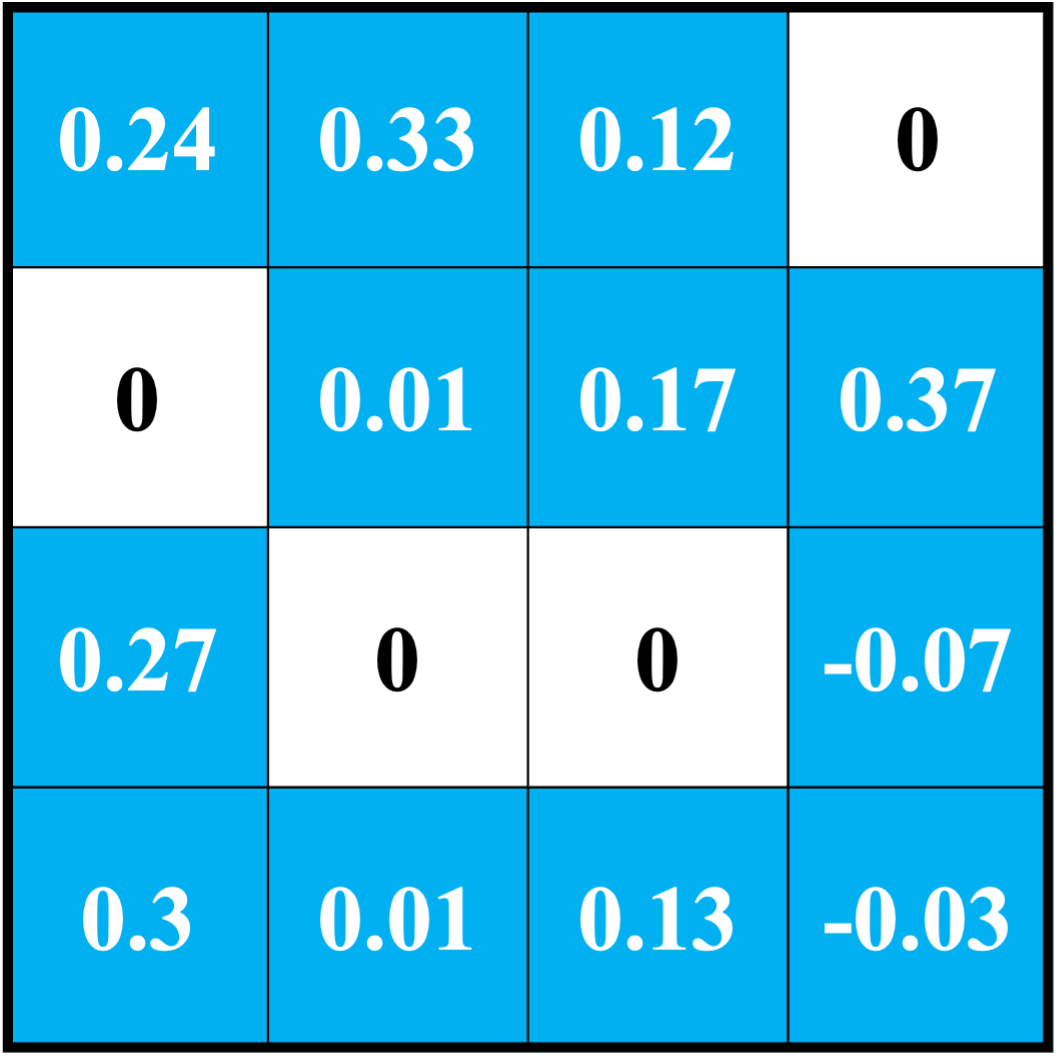}
\label{fig:activation_80}
}
\subfigure[ASH (50\%)]{
\includegraphics[width=0.165\linewidth]{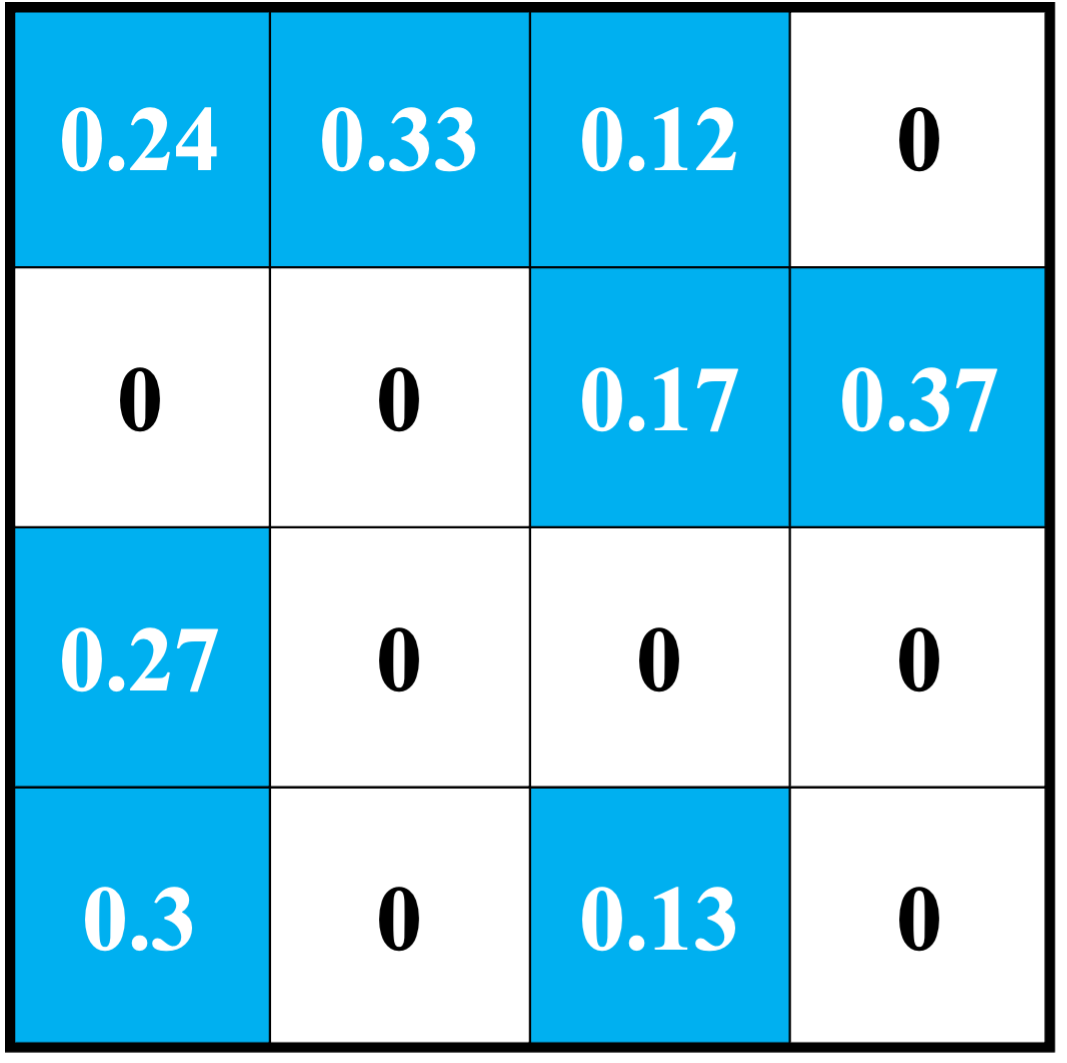}
\label{fig:activation_50}
}
\subfigure[ASH (30\%)]{
\includegraphics[width=0.165\linewidth]{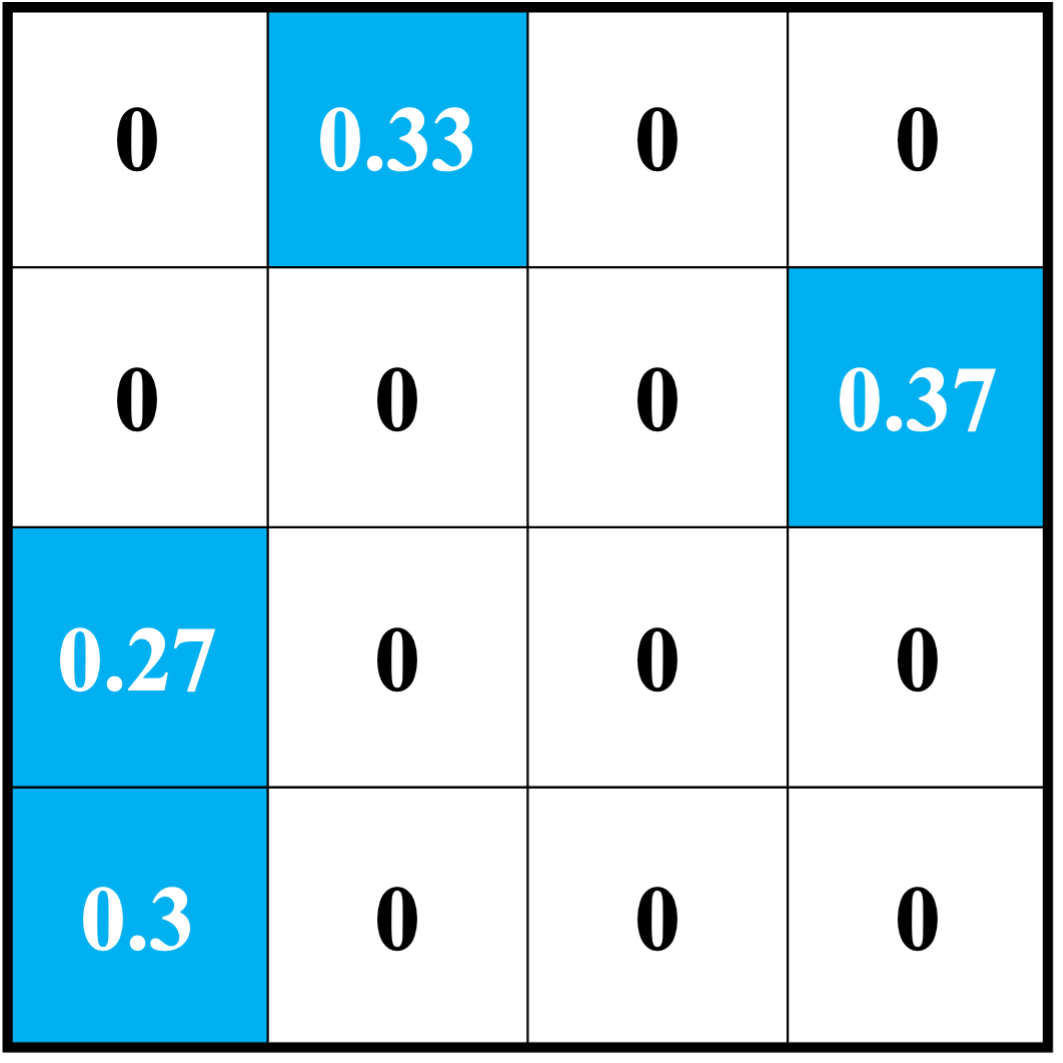}
\label{fig:activation_30}
}
\caption{Input feature-map and outputs by activation functions. ASH ($k$\%) indicates that ASH activation function samples top-$k$\% elements from input feature-map. The sampled elements by the activation functions are colored as \textcolor{customblue}{blue}.}

\end{figure}

In summary, the activation function is designed to sample the top-$k$ percentile from the input, where criteria are the values of elements. Sampling examples are presented in Fig. \ref{fig:activations}, compared to ReLU.

\section{Method}

This study aims to design an activation function for a stochastic sampling of the top-$k$ percentile elements from inputs. However, sorting or sampling methods such as a quick selection requires high computational costs. In contrast, sampling the top-$k$ percentile can be realized using a Z-score-based method in a simple calculation despite the prerequisites of a normal distribution (also known as Gaussian distribution). As discussed below, the outputs of neurons are normally distributed. Therefore, we employed the stochastic sampling to design ASH activation function.

In this section, we (1) demonstrate that the outputs of neurons are normally distributed, (2) construct a model for stochastic sampling using a Z-score, (3) formulate ASH activation function, (4) verify that ASH is parametric and trainable, and (5) search for general applications of ASH activation function.

\subsection{Background}

\paragraph{Gaussian distribution}
Many deep learning models employ normalization methods to improve their stability in training~\citep{ioffe2015batch, ulyanov2016instance, wu2018group}. In a previous study, \citet{ioffe2015batch} reported that the output of the convolutional layer, $x=Wu+b$, is more likely to have a symmetric, non-sparse distribution, that is "more Gaussian". Since most deep learning models are based on convolutional operations, the outputs of neurons are supposed to be normally distributed (Gaussian distribution). Therefore, it is concluded that the inputs of the activation functions are normally distributed when activation functions follow convolutional layers, such that $x \sim N(\mu_x, \sigma_x^2)$, where $x$ is an input feature-map of an activation function, $\mu_x$ and $\sigma_x$ are mean and standard deviation of $x$, respectively. Therefore, we can obtain the following proposition.

\textbf{Proposition 1.} The outputs of neurons in convolutional neural networks are normally distributed.

\begin{figure}[h]
  \centering
  \includegraphics[width=0.70\linewidth]{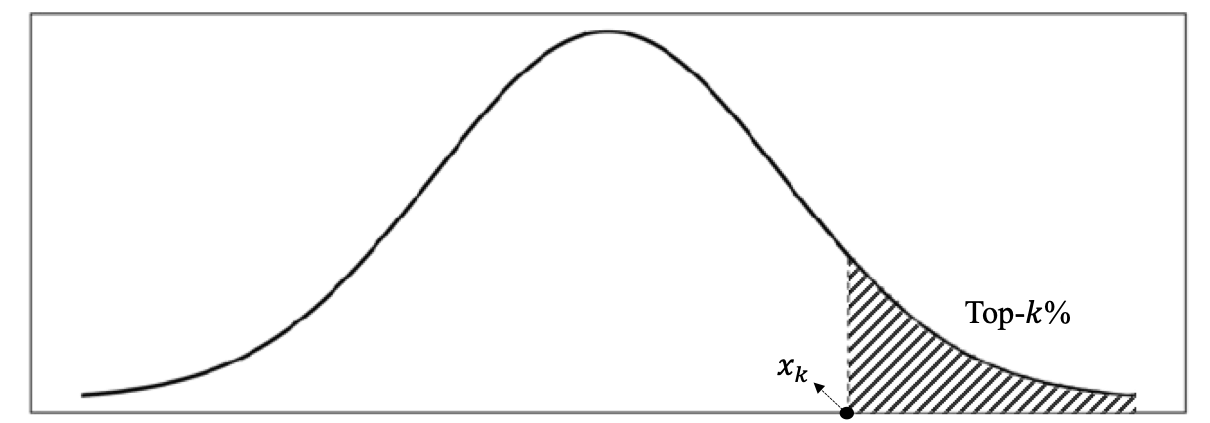}
  \caption{Top-$k$\% sampling from normal distribution}
  \label{fig:gaussian_dash}
\end{figure}

\paragraph{Sampling from a normal distribution}
This study aimed to sample the top-$k$\% elements that exhibit large values in an input feature-map. Since the area under the normal distribution indicates the percentile, sampling the top-$k$\% from a normal distribution is theoretically identical to the statement calculating the area under the curve presented in Fig. \ref{fig:gaussian_dash}. Let $F$ be a tensor, a multi-dimensional array or a matrix. $F$ should be then normally distributed, and thus we can sample the elements ($f^{(i)}$) in the top-$k$\% from $F$ using the following equation:

\begin{equation}
\label{eq:condition_norm_dist}
  P(f^{(i)} \geq z'_k) = k, \;\; k \in [0, 1] \;\; z'_k \in [-\infty, \infty]
\end{equation}
However, heavy computational costs are incurred to obtain a trivial solution from the probability density function of the normal distribution, defined as $\frac{1}{{\sigma_F \sqrt {2\pi } }}e^{{{ - \left( {x - \mu_F } \right)^2 } \mathord{\left/ {\vphantom {{ - \left( {x - \mu_F } \right)^2 } {2\sigma_F ^2 }}} \right. \kern-\nulldelimiterspace} {2\sigma_F ^2 }}}$. Therefore, we leaned to \textit{probability theory} to simplify the computation rather than \textit{calculus}. To this end, we employed the standard normal distribution (Z-score normalization) for Equation (\ref{eq:condition_norm_dist}), and we obtained the following equation:

\begin{equation}
  P(Z^{(i)} \geq z_k) = k, \;\; Z^{(i)} = \frac{f^{(i)} - \mu_F}{\sigma_F} \;\; \textit{s.t.} \;\; Z^{(i)} \sim N(0, 1)
\end{equation}

where $z_k = (z'_k-\mu_F)/\sigma_F$, which is the Z-normalized value from $z'_k$, and thus $z_k$ is subjected to $k$, indicating percentile to sample, in terms of Z-table~\citep{larsen2005introduction}. Then, we go Z-table and easily find the proper value for $z_k$, intuitively. Therefore, the condition, $Z^{(i)} = (f^{(i)}-\mu_F)/\sigma_F \geq z_k \Leftrightarrow f{(i)} \geq \mu_F + z_k\sigma_F$, is mathematically identical to sample the top-$k$\% elements from $F$. To summarize, we obtained the following proposition.

\textbf{Proposition 2.} Element $x^{(i)} \in X$, which is normally distributed, is in the top-$k$\% of $X$ if $x^{(i)} \geq \mu_X + z_k\sigma_X$, where $z_k$ is a z-value subjected to $k$ in Z-table~\citep{larsen2005introduction}.

\paragraph{Differentiation}
The mechanisms of convolutional neural networks (CNNs) have been studied in many previous works~\citep{rumelhart1986learning, bottou2010large, zhang2016derivation, hu2018improving}. Training CNN models is subjected to the backpropagation derived from the partial derivatives of loss functions by the individual convolutional parameters. Let $L$ be a loss function for the deep learning model $M$ and let $W$ be one of the variables in $M$. Then, the derivative of $L$ in terms of $W$ is represented as $\frac{\partial L}{\partial W}$, and $W$ is updated as $W \leftarrow W -\eta\frac{\partial L}{\partial W}$ with a learning rate $\eta$. On the other hand, suppose variable $\theta$ be the threshold, the function $f(x)$ is $\alpha x$ if $x \geq \theta$, otherwise 0. Thus, the partial derivative of $f$ is represented as follows:

\begin{equation}
  \frac{\partial f}{\partial x} = \begin{cases}
  \alpha, \ \text{if} \ x \geq \theta \\
  0, \ \text{otherwise}
  \end{cases}
  ,\;\; \frac{\partial f}{\partial \alpha} = \begin{cases}
  x, \ \text{if} \ x \geq \theta \\
  0, \ \text{otherwise}
  \end{cases}
  ,\;\; \frac{\partial f}{\partial \theta} = 0
\end{equation}

Here, $\alpha$ is arithmetically combined with $f(x)$, whereas $\theta$ does not. Therefore, $\alpha$ is trainable, but $\theta$ is not trainable in this context. In basic \textit{calculus}, it is trivial that if the loss function is not dependent on the variable, the partial derivative is zero, and thus the variable cannot be trained or optimized; In other words, it is invariant. Therefore, we obtain the following lemma:

\textbf{Lemma 1.} The derivative of a variable in the conditional statement is zero, and thus that the variable cannot be optimized.

\subsection{ASH Activation Function}
Let $X$ be an input of ASH activation function ($\textit{\AE}$) and be a tensor of which elements are normally distributed. Furthermore, let $x^{(i)} \in X$ be the $i$-th element in $X$. Then, by \textbf{Proposition 2}, ASH activation function, which samples the top-$k$\% elements from the input, is represented as follows:
\begin{equation}
\label{eq:ash01}
  \textit{\AE}(x^{(i)}) =
  \begin{cases}
  x^{(i)} & \text{if}\ x^{(i)} \geq \mu_X + z_k\sigma_X, \\
  0 & \text{otherwise}
  \end{cases}
\end{equation}
where $\mu_X$ and $\sigma_X$ are the mean and the standard deviation of all elements in $X$, respectively, and $z_k$ is the Z-score concerning percentile ($k$) to sample (i.e., $z=1.96$ if $k=2.5$\%, see Z-table). Equation (\ref{eq:ash01}) exhibits that ASH activation function is represented in a simple yet effective form with low computational costs.

Intuitively, we assumed that the activation level (percentile) is supposed to be different by each neuron and the tasks of the deep learning model, similar to human neurons. However, in Equation (\ref{eq:ash01}), the condition ($x^{(i)} \geq \mu_X + z_k\sigma_X$) is invariant by \textbf{Lemma 1}. Note that, $\mu_X + z_k\sigma_X$ is variable and changeable with respect to $X$, but the sampled portion ($k$\% related to $z_k$) from $X$ is invariant. Therefore, ASH in Equation (\ref{eq:ash01}) is not parametric and trainable. To make ASH be trainable and parametric, let Equation (\ref{eq:ash01}) be substituted using a proxy function as $\textit{\AE}(x^{(i)}) = x^{(i)}f(x^{(i)})$, such that the proxy function $f(x^{(i)})$ is represented as follows:
\begin{equation}
\label{eq:ash02}
  f(x^{(i)}) =
  \begin{cases}
  1 & \text{if}\ x^{(i)} - \mu_X - z_k\sigma_X \geq 0, \\
  0 & \text{otherwise}
  \end{cases}
\end{equation}

For simplicity, suppose that a Heaviside step function~\citep{weisstein2002heaviside} is defined as $H(x)=\frac{d}{dx}\text{max}(0, x)$, and thus $f(x^{(i)}) = H(x^{(i)} - \mu_X - z_k\sigma_X)$. Then, we obtain the arithmetical form to formulate ASH activation function as follows:

\begin{equation}
\label{eq:ash03}
  \textit{\AE}(x^{(i)}) = x^{(i)}H(x^{(i)} - \mu_X - z_k\sigma_X)
\end{equation}

Even with the arithmetic formula, ASH activation function in Equation (\ref{eq:ash03}) is still independent of $z_k$, and thus the $z_k$ is still not trainable. However, it is well known that the Heaviside step function is analytically approximated as $2H(x)=1+1\tanh(\alpha x)$ with a large value of $\alpha$~\citep{ILIEV2017223}, and thus ASH activation function is approximated using the smooth Heaviside step function as follows:
\begin{equation}
\label{eq:ash04}
\begin{aligned}
  \textit{\AE}(x^{(i)})  &= x^{(i)}H(x^{(i)} - \mu_X - z_k\sigma_X) \\
              &= \frac{1}{2}x^{(i)} + \frac{1}{2}x^{(i)}\tanh(\alpha (x^{(i)} - \mu_X - z_k\sigma_X)) \\
              &= \frac{x^{(i)}}{1 + e^{-2\alpha(x^{(i)} - \mu_X - z_k\sigma_X)}}
\end{aligned}
\end{equation}

Since $z_k$ is arithmetically placed, $z_k$ representing a sampling percentile is trainable, and thus ASH activation function is also trainable and parametric. By optimizing $z_k$, ASH activation functions exhibit different thresholds. Therefore, it is concluded that ASH exhibits different activation levels based on the stochastic sampling of inputs and different thresholds, similar to human neurons, synapses, and their potentials. As human neurons, the mechanism of ASH can be summarized as follows:

\paragraph{}
(1) In the training phase, each ASH activation function optimizes its $z_k$ and fine-tunes the threshold for the sampling percentile  of inputs. Thus, it implies that ASH activation function realizes the arbitrary threshold potentials as human neurons~\citep{clevert2015fast, evans2018synaptic}. Some examples of $z_k$ related to Equation (\ref{eq:ash01}) are:

\textbf{Example 1.} A small value of $z_k$, even a small negative value, implies the dense activation, and the dying ReLU problem can be solved.

\textbf{Example 2.} A large value of $z_k$ implies the sparse activation, and the sparsity can be leveraged.

\paragraph{}
(2) In the training or inference phase, ASH activation function rectifies the inputs using the learned threshold value and contexts of inputs. In particular, to sample the top-$k$ percentile, ASH employs the mean and the standard deviation of inputs, representing the contexts of the entire inputs. Therefore, it implies that ASH realizes the adaptive activation considering the contexts of inputs. Some examples of the adaptive activation related to Equation (\ref{eq:ash01}) are:

\textbf{Example 3.} A small threshold value ($\theta_s$) is employed to calculate the input $X$ that exhibits large mean and standard deviation values ($X > \theta_s$).

\textbf{Example 4.} A large threshold value ($\theta_l$) is employed to calculate the input $X'$ that exhibits large mean and standard deviation values ($X' > \theta_l \gg \theta_s$).

\subsection{Generalized Activation Function}




We found the innovation while representing Equation (\ref{eq:ash04}) using the sigmoid function $S(x) = \frac{1}{1+e^{-x}}$ as follows:

\begin{equation}
\label{eq:ash05}
\begin{aligned}
  \textit{\AE}(x^{(i)})  &= x^{(i)}S\big(-2\alpha(x^{(i)} - \mu_X - z_k\sigma_X)\big) \\
              &= x^{(i)}S(ax^{(i)}+b))
\end{aligned}
\end{equation}

In a previous work, \citet{ramachandran2017searching} introduced the leverage of automatic search techniques to discover the best performance activation function. The experiments empirically discovered that the Swish activation function is the best performance activation function, defined as $xS(x)$~\citep{ramachandran2017searching}. Intuitively, the definition of the Swish activation function is the same with Equation (\ref{eq:ash05}), and Equation (\ref{eq:ash05}) represents more generalized formula. Therefore, ASH (Adaptive SwisH) activation function provides the theoretical explanations for why Swish was the best performance activation function in the empirical evaluations. Therefore, we obtain the following.

\textbf{Lemma 3.} ASH activation function exhibits general formula for the Swish activation function.

Interestingly, the activation function designed for stochastic adaptive sampling is converged to the generalized Swish activation function. The extreme impression is that the stochastic percentile sampling by the activation function that mimics real neurons expresses the general formula of the swish activation function formerly known as state-of-the-art. Therefore, the stochastic percentile sampling can partially be applied to the Swish activation function. Additionally, it can be supposed that the Swish activation function achieved superior performance in the previous studies based on the utilization of stochastic percentile sampling.

This paper initially considered an activation function that enables stochastic percentile sampling in a mathematically effective manner. However, we found that the mathematical expression of ASH supports the theoretical background of the Swish activation function. Therefore, this paper provides the theoretical backgrounds and rationales for the Swish activation function, which was empirically investigated. It is a significant innovation to provide the mathematical theorem that the Swish activation function is derived from a stochastically designed activation function.

\section{Main Result}
Similar to a previous study~\citep{ramachandran2017searching}, we compared ASH to several baseline activation functions on various models for different tasks using public datasets. Because many activation functions have been developed, we employed some of the most commonly used activation functions, namely ReLU, leaky ReLU (LReLU)~\citep{maas2013rectifier}, parametric ReLU (PReLU)~\citep{he2015delving}, Softplus~\citep{nair2010rectified}, ELU~\citep{clevert2015fast}, SELU~\citep{klambauer2017self}, and GELU~\citep{hendrycks2016gaussian}. In our experiments, every hyper-arameter in ASH and the other activation functions was set to be the same to demonstrate the advantages of ASH compared to other activation functions. In the tables, the highest accuracy values are highlighted in \textbf{bold}.

\subsection{Classification Task}
We first compared ASH to all the baseline activation functions on the CIFAR-10, CIFAR-100 datasets, and ImageNet~\citep{russakovsky2015imagenet} datasets. We employed environments from a previous study~\citep{ramachandran2017searching} and re-implemented the baseline models of ResNet-164~\citep{he2016identity}, wide ResNet28-10~\citep{zagoruyko2016wide}, and DenseNet-100-12~\citep{huang2017densely}. Based on these different environments, small differences were reported previously, but we believe that the accuracy trends are similar. We first evaluated ASH activation function against other activation functions using the ImageNet 2012 classification dataset because ImageNet is a widely utilized dataset in classification tasks. We then evaluated all activation functions using the CIFAR-10 and CIFAR-100 datasets, which have been widely utilized as benchmarks.

\begin{table}[h]
\centering
\label{tab:my-table}
{\scriptsize
\begin{tabular}{c|ccc|ccc}
\hline\hline
\textbf{Model} & \multicolumn{3}{c|}{\textbf{Top-1 Acc. (\%)}} & \multicolumn{3}{c}{\textbf{Top-5 Acc. (\%)}} \\
\hline\hline
ReLU           & 76.4   $\pm$ 0.09           & 75.6 $\pm$ 0.10            & 77.1 $\pm$ 0.11             & 91.2 $\pm$ 0.09            & 90.7 $\pm$ 0.06            & 90.7 $\pm$ 0.06            \\
LReLU          & 77.6 $\pm$ 0.07             & 78.0 $\pm$ 0.03            & 76.6 $\pm$ 0.07             & 91.6 $\pm$ 0.10            & 91.2 $\pm$ 0.07            & 92.3 $\pm$ 0.07            \\
PLeLU          & 77.0 $\pm$ 0.13             & 78.7 $\pm$ 0.03            & 78.0 $\pm$ 0.09             & 92.9 $\pm$ 0.03            & 92.3 $\pm$ 0.14            & 92.2 $\pm$ 0.12            \\
Softplus       & 76.8 $\pm$ 0.11             & 77.3 $\pm$ 0.03            & 76.0 $\pm$ 0.05             & 91.5 $\pm$ 0.12            & 93.7 $\pm$ 0.05            & 93.8 $\pm$ 0.11            \\
ELU            & 71.6 $\pm$ 0.09             & 73.7 $\pm$ 0.09            & 74.9 $\pm$ 0.10             & 85.6 $\pm$ 0.06            & 89.8 $\pm$ 0.13            & 90.2 $\pm$ 0.08            \\
SELU           & 75.4 $\pm$ 0.13             & 78.1 $\pm$ 0.14            & 76.8 $\pm$ 0.06             & 91.6 $\pm$ 0.09            & 93.5 $\pm$ 0.10            & 90.9 $\pm$ 0.04            \\
GELU           & 76.8 $\pm$ 0.13             & 77.9 $\pm$ 0.05            & 78.0 $\pm$ 0.12             & 90.2 $\pm$ 0.11            & 92.6 $\pm$ 0.04            & 91.9 $\pm$ 0.09            \\
Swish          & 77.5 $\pm$ 0.07             & 76.6 $\pm$ 0.06            & 76.5 $\pm$ 0.05             & 92.2 $\pm$ 0.12            & 90.9 $\pm$ 0.07            & 92.2 $\pm$ 0.07            \\ \hline
ASH            & \textbf{78.5 $ \pm $ 0.06} & \textbf{78.6$ \pm $ 0.07} & \textbf{78.7$ \pm $ 0.10} & \textbf{94.0$ \pm $ 0.08} & \textbf{94.7$ \pm $ 0.07} & \textbf{94.1$ \pm $ 0.08} \\ \hline
\hline
\multicolumn{7}{c}{Table 1. ImageNet dataset. Three models are averaged. The values are mean and 95\% confidence Interval (C.I.)} \\
\end{tabular}
}
\end{table}

\begin{table}[h]
{\scriptsize
\begin{tabular}{cccc}
\hline\hline
\textbf{Model} & \textbf{ResNet} & \textbf{WRN} & \textbf{DenseNet} \\ \hline
ReLU     & 94.4   $\pm$ 0.04        & 95.6 $\pm$ 0.03          & 95.7 $\pm$ 0.02          \\
LReLU    & 94.5 $\pm$ 0.05          & 95.6 $\pm$ 0.04          & 94.7 $\pm$ 0.09          \\
PLeLU    & 94.7 $\pm$ 0.08          & 95.4 $\pm$ 0.03          & 95.1 $\pm$ 0.08          \\
Softplus & 94.3 $\pm$ 0.10          & 94.2 $\pm$ 0.08          & 95.2 $\pm$ 0.07          \\
ELU      & 93.5 $\pm$ 0.10          & 93.8 $\pm$ 0.09          & 94.5 $\pm$ 0.11          \\
SELU     & 94.5 $\pm$ 0.05          & 95.8 $\pm$ 0.07          & 94.9 $\pm$ 0.10       \\
GELU     & 95.2 $\pm$ 0.04          & 95.7 $\pm$ 0.06          & 94.8 $\pm$ 0.10          \\
Swish    & 95.5 $\pm$ 0.09          & 95.6 $\pm$ 0.08          & 95.2 $\pm$ 0.03          \\ \hline
ASH      & \textbf{95.7} $\pm$ 0.08 & \textbf{96.7} $\pm$ 0.04 & \textbf{96.0} $\pm$ 0.11 \\
\hline\hline
\multicolumn{4}{c}{Table 2. CIFAR-10 with mean values and 95\% C.I.} \\
\end{tabular}
\hfill
\begin{tabular}{cccc}
\hline\hline
\textbf{Model} & \textbf{ResNet} & \textbf{WRN} & \textbf{DenseNet} \\ \hline
ReLU     & 74.5 $\pm$ 0.10          & 78.4 $\pm$ 0.04          & 84.0 $\pm$ 0.09           \\
LReLU    & 75.3 $\pm$ 0.06          & 77.9 $\pm$ 0.07          & 82.2 $\pm$ 0.07         \\
PLeLU    & 74.7 $\pm$ 0.06          & 77.6 $\pm$ 0.06          & 82.1 $\pm$ 0.07         \\
Softplus & 76.1 $\pm$ 0.05          & 78.6 $\pm$ 0.06          & 84.1 $\pm$ 0.02         \\
ELU      & 75.0 $\pm$ 0.08          & 76.4 $\pm$ 0.09          & 80.8 $\pm$ 0.05         \\
SELU     & 73.4 $\pm$ 0.05          & 74.4 $\pm$ 0.08          & 81.4 $\pm$ 0.06         \\
GELU     & 75.0 $\pm$ 0.05          & 78.2 $\pm$ 0.10          & 84.0 $\pm$ 0.02           \\
Swish    & 75.7 $\pm$ 0.10          & 78.9 $\pm$ 0.05          & 84.0 $\pm$ 0.03           \\ \hline
ASH      & \textbf{76.5} $\pm$ 0.08 & \textbf{79.2} $\pm$ 0.06 & \textbf{84.6} $\pm$ 0.06 \\ \hline\hline
\multicolumn{4}{c}{Table 3. CIFAR-100 with mean values and 95\% C.I.} \\
\end{tabular}
}
\end{table}

The ImageNet dataset evaluations were averaged based on the accuracy values of the three deep learning models. The results in Tables 1 to 3 highlight the outstanding performance of ASH activation function in terms of improving predictive accuracy. Because deep learning models for classification tasks demand sparsity, it is intuitive that ASH activation function improves accuracy compared to other activation functions. 
 
\subsection{Detection Task}
We compared ASH to all the baseline activation functions on the COCO~\citep{lin2014microsoft} and PASCAL VOC~\citep{everingham2010pascal} datasets for the detection task. Both of these datasets are widely used as benchmarks for detection tasks. We employed the same environments as the classification task and implemented the baseline models of Mask-R-CNN~\citep{he2017mask}, SSD~\citep{liu2016ssd}, and YOLOv4~\citep{bochkovskiy2020yolov4}. For detection tasks, deep learning models output bounding boxes representing the locations of target objects. We exploited mAP@50 as an evaluation metric based on its popularity for detection tasks.

\begin{table}[h]
{\scriptsize
\begin{tabular}{cccc}
\hline\hline
\textbf{Model} & \textbf{MR-CNN} & \textbf{SSD} & \textbf{YOLOv5} \\ \hline
ReLU     & 68.5 $\pm$ 0.02          & 70.1 $\pm$ 0.07           & 72.1 $\pm$ 0.04          \\
LReLU    & 68.9 $\pm$ 0.06          & 70.6 $\pm$ 0.10           & 72.6 $\pm$ 0.03          \\
PLeLU    & 69.4 $\pm$ 0.10          & 71.0 $\pm$ 0.11           & 73.1 $\pm$ 0.11          \\
Softplus & 69.4 $\pm$ 0.03          & 71.0 $\pm$ 0.08           & 73.0 $\pm$ 0.09            \\
ELU      & 69.4 $\pm$ 0.10          & 71.1 $\pm$ 0.10           & 73.0 $\pm$ 0.06            \\
SELU     & 69.7 $\pm$ 0.06          & 71.2 $\pm$ 0.04           & 73.3 $\pm$ 0.10          \\
GELU     & 70.0 $\pm$ 0.08          & 71.6 $\pm$ 0.05           & 73.7 $\pm$ 0.04          \\
Swish    & 70.4 $\pm$ 0.08          & 72.0 $\pm$ 0.03           & 74.0 $\pm$ 0.08            \\ \hline
ASH      & \textbf{71.1 $\pm$ 0.06} & \textbf{72.7 $\pm$ 0.02} & \textbf{74.8 $\pm$ 0.09}   \\
\hline\hline\\
\multicolumn{4}{c}{Table 4. COCO with mean values and 95\% C.I.}
\end{tabular}
\hfill
\begin{tabular}{cccc}
\hline\hline
\textbf{Model} & \textbf{MR-CNN} & \textbf{SSD} & \textbf{YOLOv5} \\ \hline
ReLU     & 65.8 $\pm$ 0.03          & 67.3 $\pm$ 0.07          & 69.3 $\pm$ 0.09          \\
LReLU    & 66.9 $\pm$ 0.07          & 68.4 $\pm$ 0.07          & 70.5 $\pm$ 0.07          \\
PLeLU    & 67.8 $\pm$ 0.03          & 69.2 $\pm$ 0.07          & 71.2 $\pm$ 0.09          \\
Softplus & 67.9 $\pm$ 0.05          & 69.4 $\pm$ 0.03          & 71.4 $\pm$ 0.03          \\
ELU      & 67.8 $\pm$ 0.05          & 69.3 $\pm$ 0.06          & 71.4 $\pm$ 0.06          \\
SELU     & 68.4 $\pm$ 0.07          & 69.9 $\pm$ 0.03          & 72.1 $\pm$ 0.11          \\
GELU     & 68.9 $\pm$ 0.07          & 70.5 $\pm$ 0.03          & 72.4 $\pm$ 0.11          \\
Swish    & 69.0 $\pm$ 0.04          & 70.6 $\pm$ 0.02          & 72.6 $\pm$ 0.08          \\ \hline
ASH      & \textbf{70.5 $\pm$ 0.06} & \textbf{72.1 $\pm$ 0.04} & \textbf{74.1 $\pm$ 0.11}   \\
\hline\hline
\\
\multicolumn{4}{c}{Table 5. PASCAL VOC with mean values and 95\% C.I.}
\end{tabular}
}
\end{table}

The quantitative results in Tables 4 and 5 highlight the outstanding performance of ASH activation function compared to other activation functions. A higher mAP indicates that the predicted bounding boxes are closer to the annotations. ASH activation function provides superior performance for detecting target objects in various datasets for various deep learning models. Because the deep learning models used for detection tasks demand locality to generate bounding boxes, it is expected that $z_k$ will be small, demonstrating that greater activation can be realized using ASH activation function compared to the models used for the classification task. 

\begin{table}[b!]
{\scriptsize
\begin{tabular}{cccc}
\hline\hline
\textbf{Model} & \textbf{U-Net} & \textbf{DLV3+} & \textbf{EfficientNet} \\ \hline
ReLU     & 49.4 $\pm$ 0.04          & 50.7 $\pm$ 0.06          & 52.2 $\pm$ 0.03          \\
LReLU    & 49.8 $\pm$ 0.07          & 51.0 $\pm$ 0.09          & 52.3 $\pm$ 0.05          \\
PLeLU    & 49.9 $\pm$ 0.02          & 51.0 $\pm$ 0.03          & 52.5 $\pm$ 0.06          \\
Softplus & 50.1 $\pm$ 0.11          & 51.2 $\pm$ 0.10          & 52.8 $\pm$ 0.03          \\
ELU      & 50.3 $\pm$ 0.04          & 51.4 $\pm$ 0.09          & 52.8 $\pm$ 0.08          \\
SELU     & 50.9 $\pm$ 0.04          & 52.1 $\pm$ 0.03          & 53.5 $\pm$ 0.06          \\
GELU     & 50.9 $\pm$ 0.07          & 52.2 $\pm$ 0.08          & 53.5 $\pm$ 0.05          \\
Swish    & 51.3 $\pm$ 0.05          & 52.4 $\pm$ 0.02          & 53.9 $\pm$ 0.07          \\ \hline
ASH      & \textbf{53.4 $\pm$ 0.05} & \textbf{54.7 $\pm$ 0.08} & \textbf{56.3 $\pm$ 0.09}         \\ 
\hline\hline
\multicolumn{4}{c}{Table 6. ADE20K with mean values and 95\% C.I.}
\end{tabular}
\hfill
\begin{tabular}{cccc}
\hline\hline
\textbf{Model} & \textbf{U-Net} & \textbf{DLV3+} & \textbf{EfficientNet} \\ \hline
ReLU     & 74.3 $\pm$ 0.05          & 76.0 $\pm$ 0.08           & 78.1 $\pm$ 0.07          \\
LReLU    & 76.2 $\pm$ 0.06          & 78.0 $\pm$ 0.05           & 80.3 $\pm$ 0.10          \\
PLeLU    & 77.0 $\pm$ 0.07          & 78.9 $\pm$ 0.06           & 81.0 $\pm$ 0.08            \\
Softplus & 77.1 $\pm$ 0.09          & 78.8 $\pm$ 0.03           & 81.2 $\pm$ 0.10           \\
ELU      & 77.2 $\pm$ 0.10          & 79.0 $\pm$ 0.05           & 81.3 $\pm$ 0.02          \\
SELU     & 78.2 $\pm$ 0.06          & 80.1 $\pm$ 0.03           & 82.4 $\pm$ 0.03          \\
GELU     & 78.9 $\pm$ 0.06          & 80.7 $\pm$ 0.05           & 83.2 $\pm$ 0.08          \\
Swish    & 78.8 $\pm$ 0.11          & 80.7 $\pm$ 0.04           & 82.9 $\pm$ 0.04          \\ \hline
ASH      & \textbf{81.2 $\pm$ 0.04} & \textbf{83.2 $\pm$ 0.04} & \textbf{85.5 $\pm$ 0.05}         \\
\hline\hline
\multicolumn{4}{c}{Table 7. PASCAL VOC with mean values and 95\% C.I.}
\end{tabular}
}
\end{table}

\subsection{Segmentation Task}
We compared ASH to all of the baseline activation functions on the ADE20K~\citep{zhou2017scene} and PASCAL VOC~\citep{everingham2010pascal} datasets for the segmentation task. Both datasets include many target objects in one scene. Therefore, they are widely utilized as benchmarks for segmentation tasks. We employed the same environments as the classification and detection tasks and implemented the baseline models of U-Net~\citep{ronneberger2015u}, DeepLabV3+(DLV3+)~\citep{chen2018encoder}, and EfficientNet~\citep{tan2019efficientnet}. Similar to other general benchmarks, we adopted intersection over union (IoU) and mean IoU (mIoU) values as evaluation metrics based on their popularity for segmentation tasks.

Similar to the previous tasks, the quantitative results in Tables 6 and 7 highlight the outstanding performance of ASH activation function compared to the other activation functions. Because locality is important for segmenting target objects from the background in segmentation tasks, it is intuitive that ASH activation function improves locality during feature extraction. The experimental results demonstrate that superior segmentation performance can be realized by using ASH activation function, which aids significantly in localizing target objects.

\subsection{Training Time}
\begin{figure}[h]
  \centering
  \includegraphics[width=0.75\linewidth]{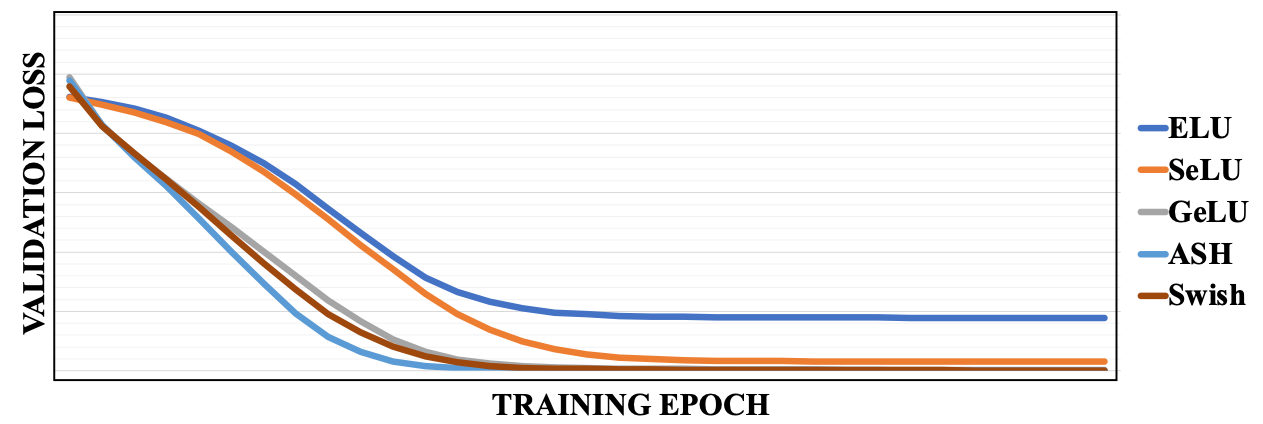}
  \caption{Validation loss values alongside the training epoch for the activation functions. The validation losses are averaged from the results of all experiments.}
  \label{fig:trainingtime}
\end{figure}

We empirically explored the effectiveness of ASH activation function in terms of training time by monitoring the validation loss values for all activation functions. Fig.~\ref{fig:trainingtime} reveals that the loss values of ASH activation function exhibit a steeper slope than those of the other activation functions. Therefore, because ASH activation function reaches convergence significantly faster than the other activation functions, we can empirically conclude that ASH has a superior effect in terms of reducing training time.

\paragraph{}
Through our experiments, we explored the outstanding performance of ASH activation function compared to other activation functions, including improvements in accuracy, sparsity, training time, and localization. In this study, the experimental results demonstrate the outstanding performance of ASH activation function. Additionally, supporting experiments and the results of other tasks such as image generation is presented in the \textit{Supplementary Material}.

\section{Conclusions}
In this paper, we proposed a novel activation function to rectify inputs using an adaptive threshold considering the entire contexts of inputs more like human neurons. To this end, we designed an activation function to extract elements in the top-$k$ percentile from the input feature-map. Since sorting algorithm-based selections or quick selection algorithm demands a heavy computational cost, we employed the stochastic technique utilizing normal distribution to realize stochastic percentile sampling. Based on the mathematical derivations, we implemented ASH activation function in simple yet effective formula ($f(x) = x\cdot\text{sigmoid}(ax+b)$) with low computational cost for sampling the top-$k$ percentile from the input. In addition, we implemented ASH activation function, realizing (1) the adaptive threshold by employing the Z-score-based trainable variables and (2) the perception of entire contexts in rectifying an input by utilizing the mean and standard deviation of the input. Meanwhile, ASH activation function represented the generalized form of the Swish activation function that was empirically searched in the previous study. Therefore, this study also exhibited a novel contribution of the mathematical proofs for the state-of-the-art performance of the Swish activation function. Experiments using various deep learning models on different tasks (classification, detection, and segmentation) demonstrated superior performance for ASH activation function, in terms of accuracy, localization, and training time.

\section*{Acknowledgment}
This work was supported in part by the National Research Foundation of Korea (NRF) under Grant NRF-2020R1A2B5B01002786 and in part by the Bio \& Medical Technology Development Program of the National Research Foundation (NRF) funded by the Korean government (MSIT) (No.2017M3A9G8084463).

\clearpage
\section*{Checklist}

The checklist follows the references. Please
read the checklist guidelines carefully for information on how to answer these
questions. For each question, change the default \answerTODO{} to \answerYes{},
\answerNo{}, or \answerNA{}. You are strongly encouraged to include a {\bf
justification to your answer}, either by referencing the appropriate section of
your paper or providing a brief inline description. For example:
\begin{itemize}
 \item Did you include the license to the code and datasets? \answerNA{}
 \item Did you include the license to the code and datasets? \answerNA{}
 \item Did you include the license to the code and datasets? \answerNA{}
\end{itemize}
Please do not modify the questions and only use the provided macros for your
answers. Note that the Checklist section does not count towards the page
limit. In your paper, please delete this instructions block and only keep the
Checklist section heading above along with the questions/answers below.

\begin{enumerate}

\item For all authors...
\begin{enumerate}
 \item Do the main claims made in the abstract and introduction accurately reflect the paper's contributions and scope?
  \answerYes{See Introduction}
 \item Did you describe the limitations of your work?
  \answerYes{See Section 2.1, Appendix}
 \item Did you discuss any potential negative societal impacts of your work?
  \answerNA{}
 \item Have you read the ethics review guidelines and ensured that your paper conforms to them?
  \answerYes{}
\end{enumerate}

\item If you are including theoretical results...
\begin{enumerate}
 \item Did you state the full set of assumptions of all theoretical results?
  \answerYes{See Section 2.1}
    \item Did you include complete proofs of all theoretical results?
  \answerYes{See Section 2}
\end{enumerate}

\item If you ran experiments...
\begin{enumerate}
 \item Did you include the code, data, and instructions needed to reproduce the main experimental results (either in the supplemental material or as a URL)?
  \answerYes{See Abstract}
 \item Did you specify all the training details (e.g., data splits, hyperparameters, how they were chosen)?
  \answerYes{See Appendix}
    \item Did you report error bars (e.g., with respect to the random seed after running experiments multiple times)?
  \answerYes{statistical analysis were provided using confidence interval}
    \item Did you include the total amount of compute and the type of resources used (e.g., type of GPUs, internal cluster, or cloud provider)?
  \answerYes{See Appendix}
\end{enumerate}

\item If you are using existing assets (e.g., code, data, models) or curating/releasing new assets...
\begin{enumerate}
 \item If your work uses existing assets, did you cite the creators?
  \answerYes{}
 \item Did you mention the license of the assets?
  \answerYes{}
 \item Did you include any new assets either in the supplemental material or as a URL?
  \answerNA{}
 \item Did you discuss whether and how consent was obtained from people whose data you're using/curating?
  \answerYes{}
 \item Did you discuss whether the data you are using/curating contains personally identifiable information or offensive content?
  \answerNA{}
\end{enumerate}

\item If you used crowdsourcing or conducted research with human subjects...
\begin{enumerate}
 \item Did you include the full text of instructions given to participants and screenshots, if applicable?
  \answerNA{}
 \item Did you describe any potential participant risks, with links to Institutional Review Board (IRB) approvals, if applicable?
  \answerNA{}
 \item Did you include the estimated hourly wage paid to participants and the total amount spent on participant compensation?
  \answerNA{}
\end{enumerate}

\end{enumerate}


\appendix


\renewcommand{\figurename}{Supplementary Figure}
\renewcommand{\theequation}{A\arabic{equation}}

\setcounter{equation}{0}
\setcounter{figure}{0}

\section*{Appendix A. Environment Description}
The server included two CPUs of Intel(R) Xeon(R) Gold 6226R CPU @ 2.90GHz, 128GB RAMs, and eight Titan-Xp GPUs. Besides, we developed a deep learning models and activation functions using Tensorflow version 1~\citep{abadi2016tensorflow} for the precise implementation. For the training, the batch size~\citep{bottou2010large} of the training was set to 32, and the Adam optimizer was utilized with the default values of all parameters~\citep{kingma2014adam}.

\section*{Appendix B. Properties of ASH}

\paragraph{}
ASH activation function is formulated as the following:

\begin{equation}
\begin{aligned}
  \textit{\AE}(x^{(i)})  &= x^{(i)}S\big(-2\alpha(x^{(i)} - \mu_X - z_k\sigma_X)\big) \\
  &= \begin{cases}
  x^{(i)} & \text{if}\ x^{(i)} \geq \mu_X + z_k\sigma_X, \\
  0 & \text{otherwise}
  \end{cases}
\end{aligned}
\end{equation}

where $x^{(i)}$ is an element in input feature map $X$, and $\mu_X$ and $\sigma_X$ is the mean and the standard deviations of all elements in $X$. $S$ indicates \textit{sigmoid} function, and $z_k$ is the variable with regard to sampling the top-$k$\% percentile from $X$. Intuitively, ASH activation function is the threshold-based activation function rectifying inputs, and we obtained the following properties:

\paragraph{}
\textbf{Property 1.} ASH activation function is parametric.

We represented ASH activation function to be arithmetic and trainable due to $z_k$ concerning sampling percentile, and thus ASH is trainable and parametric. Thus, ASH activation function could exhibit different thresholds concerning the location or depth in a network. ASH activation function in the early layer exhibits a small threshold (large percentile) to retain substantial information, whereas ASH in deeper layers exhibits a small comparative percentile to rectify futile information. This property improves the superior rectification of ASH in deep neural networks.

\paragraph{}
\textbf{Property 2.} ASH activation function provides output concerning the contexts of the input.

Since the threshold value ($\mu_X + z_k\sigma_X$) is concerning the distribution of input $X$, the threshold value could be further fine-tuned with regard to the inputs. Compared to other threshold-based activation functions, ASH exploits an adaptive threshold value, and thus it exhibits superior accuracy regardless of the variations in datasets.

\paragraph{}
Due to the novel properties, ASH activation function exhibits an improvement in imitating human neurons. More like human neurons compared to other activation functions, ASH provides output regarding the contexts of an input feature-map, and ASH exhibits different threshold values regarding the location, depth, or the types of the connected layers. To summarize, ASH exhibits novelty in imitating human neurons in terms of the activation function.

\clearpage

\section*{Appendix C. Training curves}

\begin{figure}[h!]
    \centering
    \includegraphics[width=1.00\linewidth]{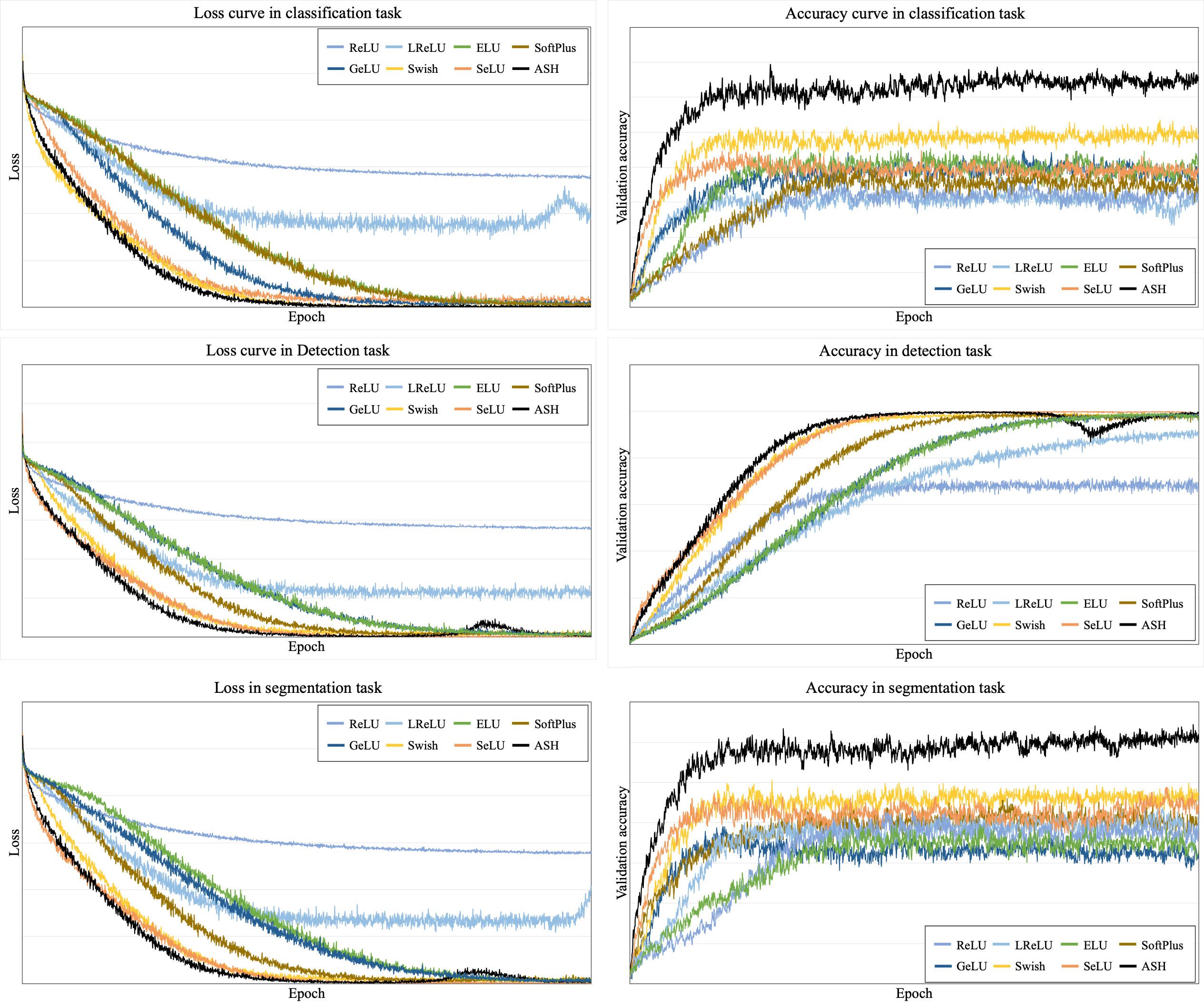}
    \caption{Average training graph of ResNet-164,Wide ResNet28-10, and DenseNet-100-12 on ImageNet dataset using various activation functions along with ASH.}
    \label{fig:training curves}
\end{figure}

Supplementary Fig. \ref{fig:training curves} illustrates the training graph of loss values and validation accuracies. The experimental results demonstrate that ASH activation function is superior in training deep learning models for various tasks, including classification, detection, and segmentation. In particular, in the classification task, since ASHs in the early layers provide broad activation and ASHs at the end of the model rectify informative features (\textbf{Property 1}), ASH significantly improves the training efficiency and accuracy. Similarly, ASH exhibits significant localization properties like attention mechanism, and thus ASH achieved superior segmentation performance. On the other hand, ASH improves the predictive accuracy of the bounding boxes in the detection task, whereas it degrades the confidence score due to its localization property. Therefore, the accuracy of ASH activation function somewhat decreases at the end of the training. Here, the x-axis indicates the percentage of training epoch, and they were averaged. In addition, the y-axis indicates the range of (0, 0.8).

\clearpage

\section*{Appendix D. Classification task}

\begin{figure}[h!]
    \centering
    \includegraphics[width=1.00\linewidth]{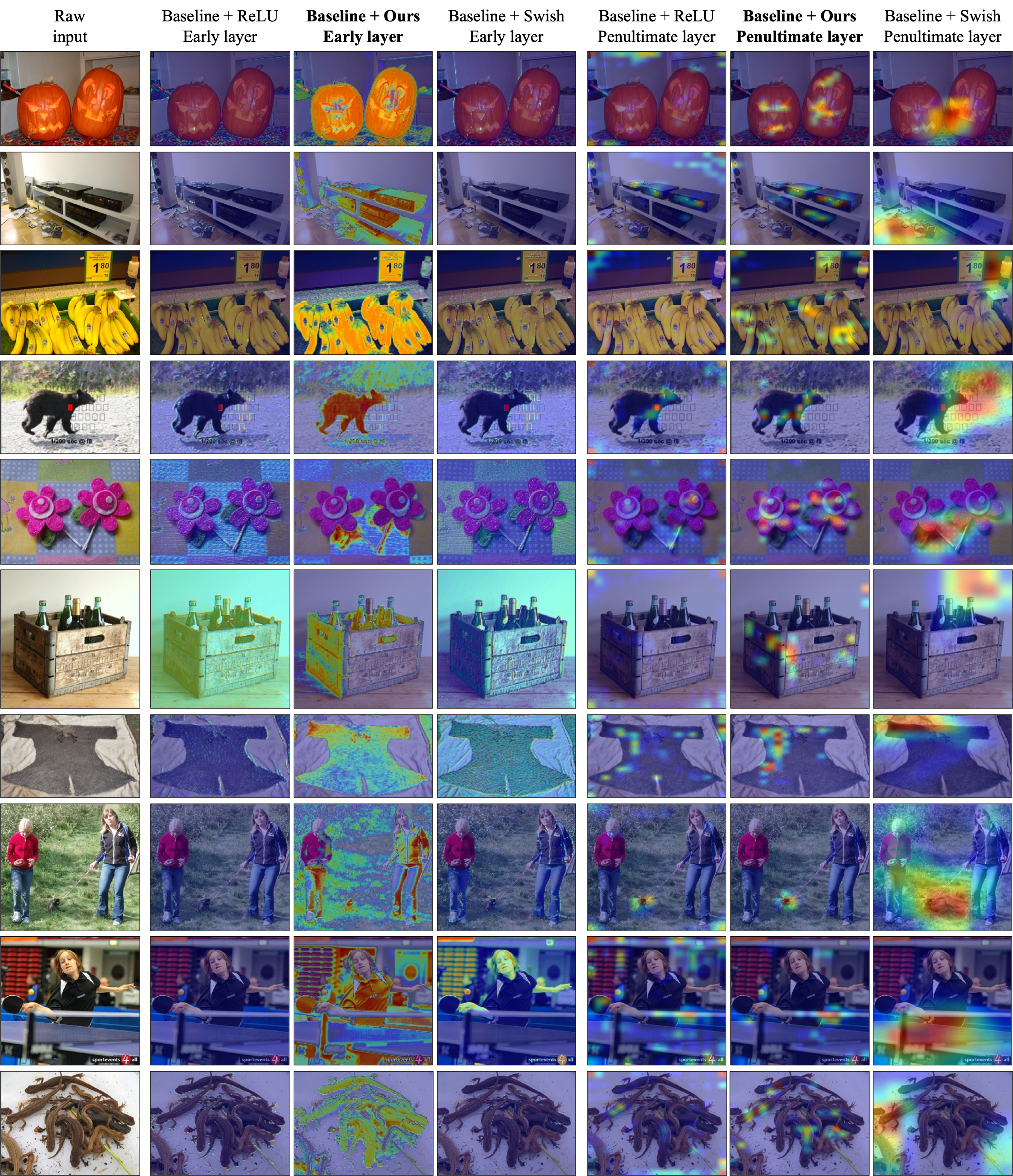}
    \caption{Grad-CAM samples generated by Baseline models with ReLU, ASH, and Swish activation functions using Imagenet dataset. ResNet-164 (1$\sim$5 rows) and Dense-Net (6$\sim$10 rows) are used as baseline models.}
    \label{fig:appendix_cam}
\end{figure}

Supplementary Fig. \ref{fig:appendix_cam} illustrates the GRAD-CAM~\citep{selvaraju2017grad} samples by using ResNet-164 and Dense-Net models with ReLU, Swish, and ASH activation function in the classification task of ImageNet dataset. Here, since ASH is based on the threshold-based activation function, ASH exhibits discrete activations like ReLU. In Supplementary Fig. \ref{fig:appendix_cam} \textbf{Property 1} is clearly illustrated. In the early layer, ASH activation function provides sufficiently broad but informative activations with regard to the target objects to forward layers. Besides, ASH at the end of the models exhibits the activations that are discrete but localized onto the target object. Therefore, ASH activation function could provide informative activations to the deep learning models, and thus it could improve the superior accuracy in every task.

\clearpage

\section*{Appendix E. Segmentation task}

\begin{figure}[h!]
    \centering
    \includegraphics[width=0.90\linewidth]{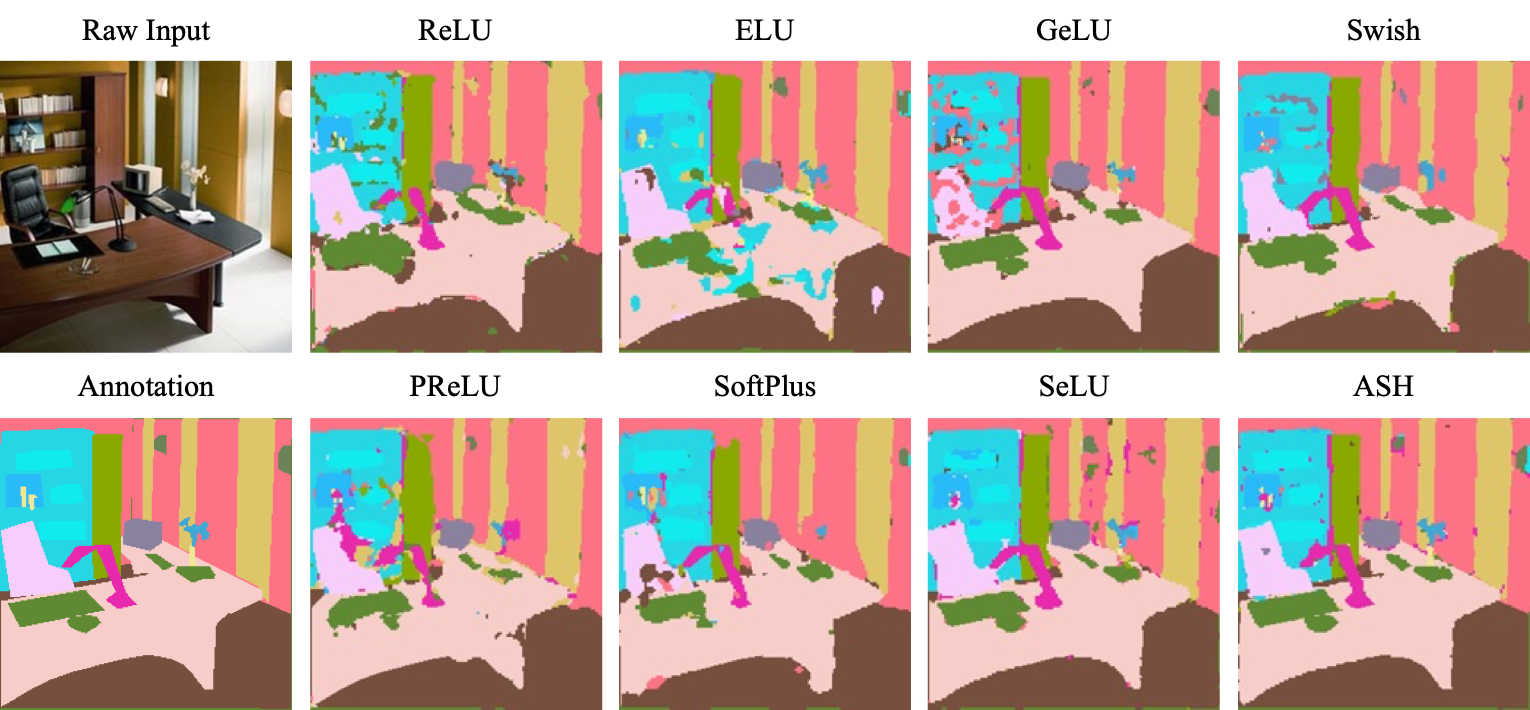}
    \caption{Samples segmented by U-Net~\citep{ronneberger2015u} with ReLU, PReLU, ELU, SoftPlus, GELU, SeLU, Swish, and ASH activation functions using ADE20K dataset.}
    \label{fig:segmentation_01}
\end{figure}

\begin{figure}[h!]
    \centering
    \includegraphics[width=0.90\linewidth]{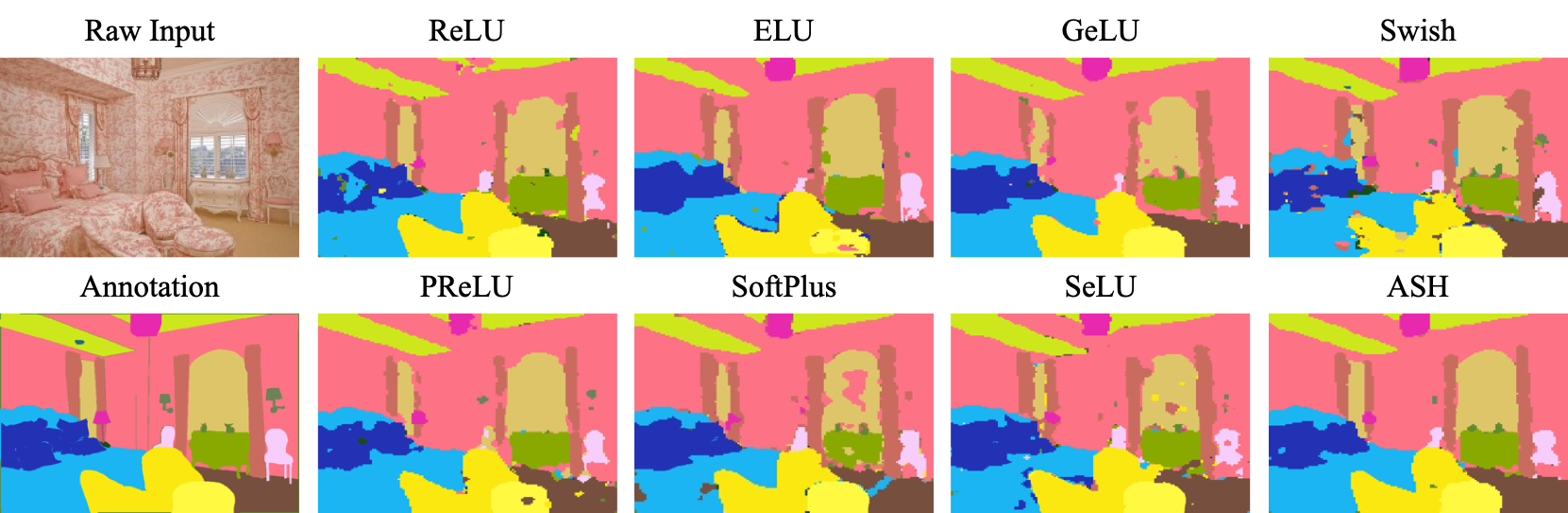}
    \caption{Samples segmented by DeepLabV3+~\citep{chen2018encoder} with ReLU, PReLU, ELU, SoftPlus, GELU, SeLU, Swish, and ASH activation functions using ADE20K dataset.}
    \label{fig:segmentation_02}
\end{figure}

\begin{figure}[h!]
    \centering
    \includegraphics[width=0.90\linewidth]{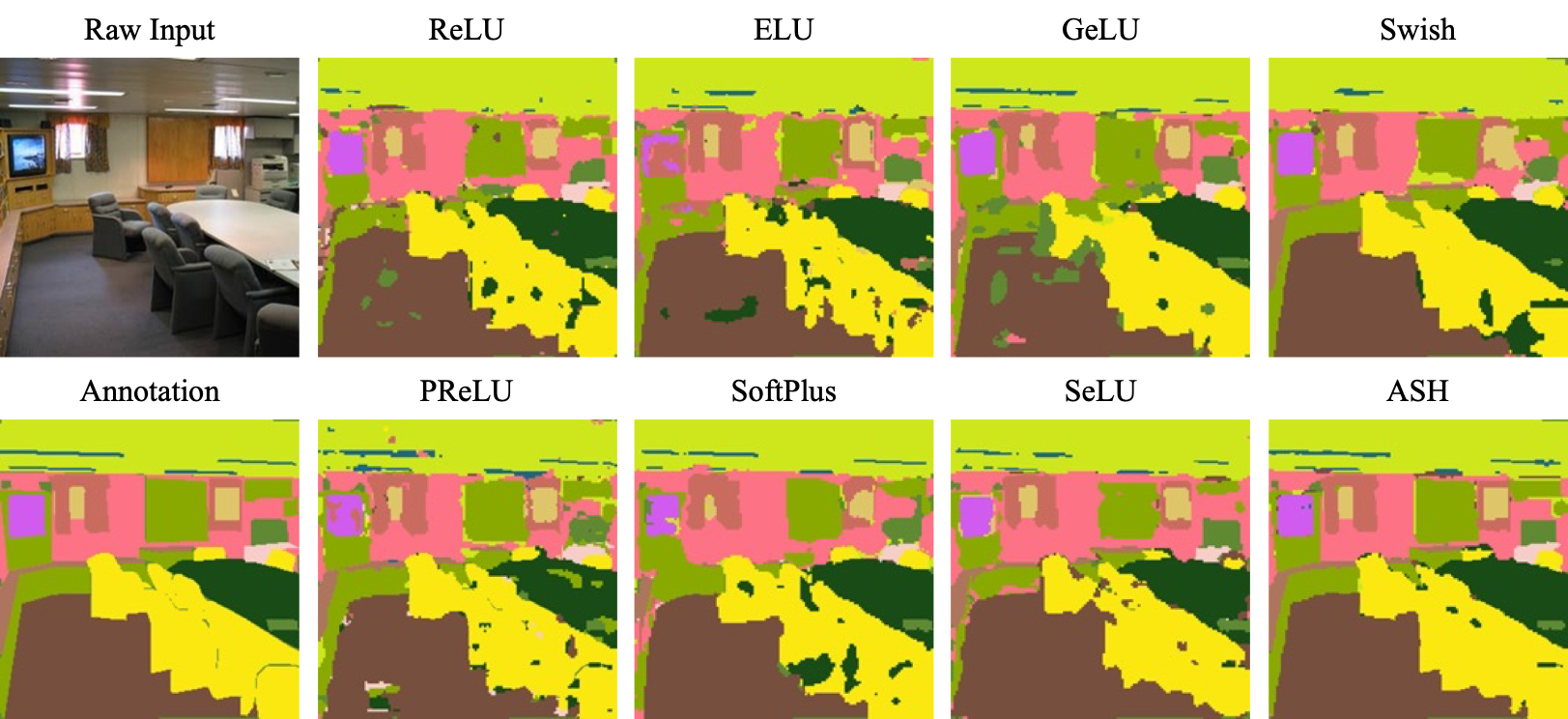}
    \caption{Samples segmented by EfficientNet~\citep{tan2019efficientnet} with ReLU, PReLU, ELU, SoftPlus, GELU, SeLU, Swish, and ASH activation functions using ADE20K dataset.}
    \label{fig:segmentation_03}
\end{figure}

\clearpage

\section*{Appendix F. Image generation task}

\begin{figure}[h!]
    \centering
    \includegraphics[width=1.00\linewidth]{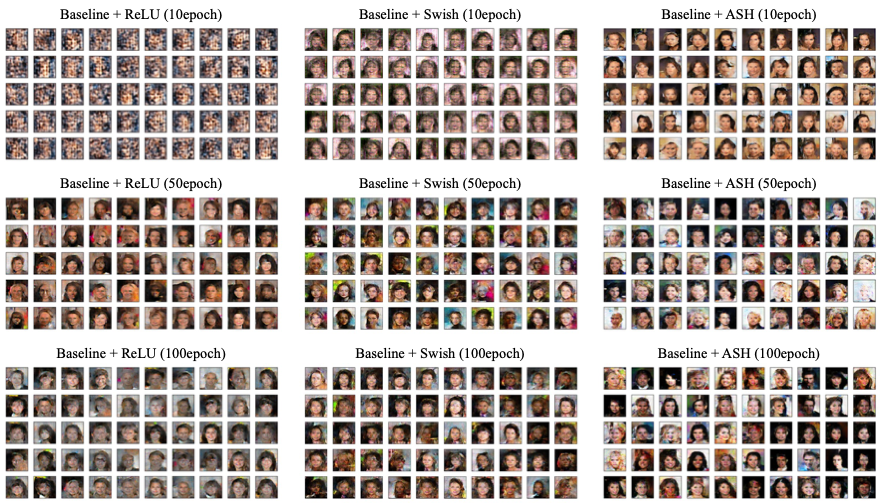}
    \caption{Samples generated by DCGAN with ReLU, Swish, and ASH activation functions using celebA dataset.}
    \label{fig:image_generation}
\end{figure}

Supplementary Fig. \ref{fig:image_generation} illustrates the generated samples by DCGAN~\citep{radford2015unsupervised} with ReLU, Swish, and ASH activation functions using celebA dataset~\citep{yang2015facial}. Despite the similar quantitative results by every activation function, ASH activation function significantly reduces the training time. In particular, the generated images by the DCGAN model with ASH activation function are explicitly exhibited as more like human from the early epoch (10). The experiment also demonstrates that ASH activation function could significantly improve the training efficiency due to its advantages of (\textbf{Property 1}) and (\textbf{Property 2}). 

\clearpage
\section*{Appendix G. Formulation of ASH}
\begin{figure}[h!]
    \centering
    \includegraphics[width=1.00\linewidth]{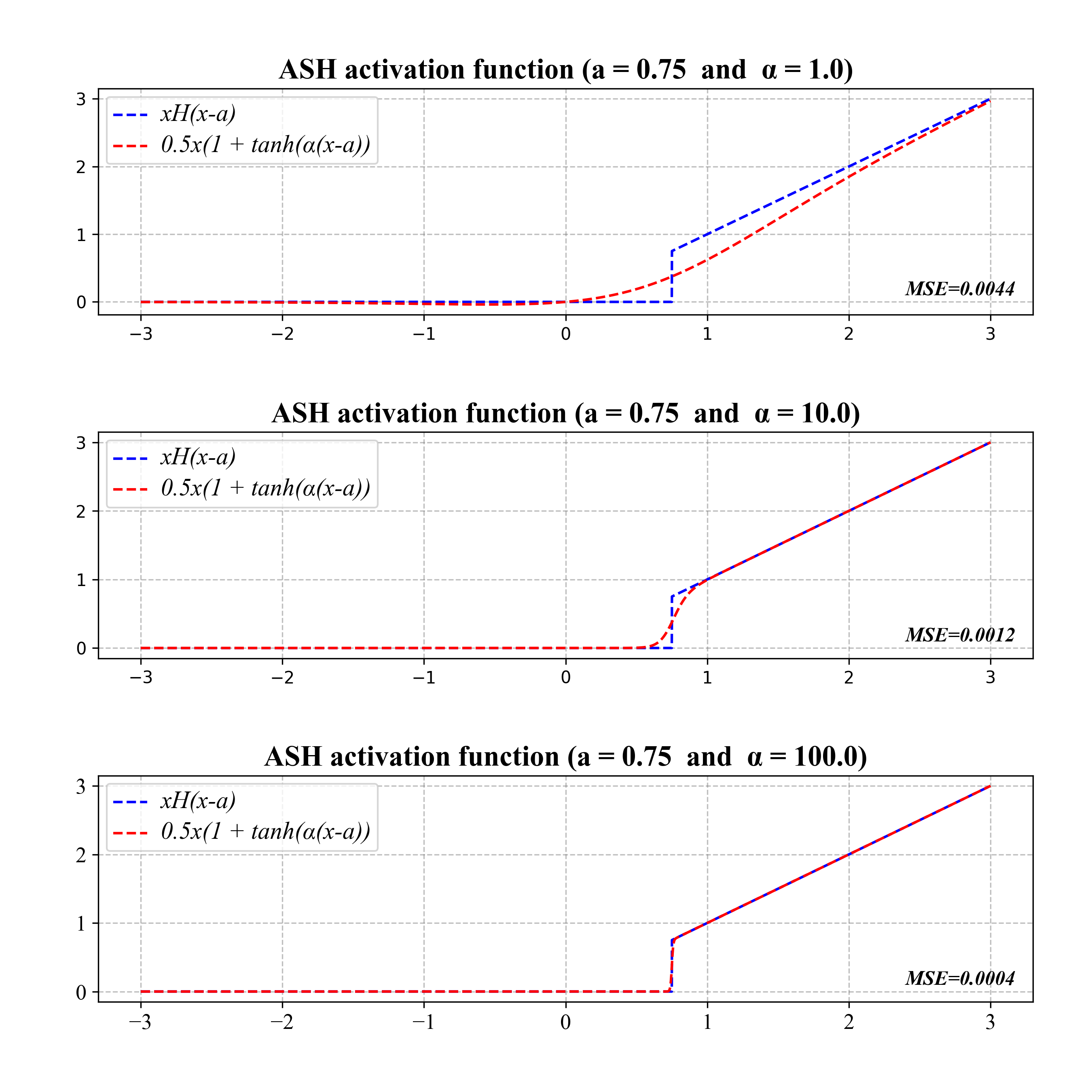}
    \caption{Various versions of ASH activation functions with various values of $\alpha$ in Equation (10). As $\alpha$ increasing, two functions are approximated as similar.}
    \label{fig:ash_versions}
\end{figure}
In equation (10), ASH activation function is initially defined using Heaviside Step Function, and it is approximated using the sigmoid function as follows:

\begin{equation}
  \textit{\AE}(x^{(i)})  = x^{(i)}H(x^{(i)} - \mu_X - z_k\sigma_X)
\end{equation}
\begin{equation}
    \textit{\AE}(x^{(i)})   = \frac{1}{2}x^{(i)} + \frac{1}{2}x^{(i)}\tanh(\alpha (x^{(i)} - \mu_X - z_k\sigma_X))
\end{equation}
\begin{equation}
     \textit{\AE}(x^{(i)})  = \frac{x^{(i)}}{1 + e^{-2\alpha(x^{(i)} - \mu_X - z_k\sigma_X)}}
\end{equation}

Here, Equations (A2) and (A3) exhibit the same equation since $\tanh (x) = \frac{e^x - e^{-x}}{e^{x} + e^{-x}}$. To simplify, Equation (A3) can be expressed using the substitution as the following:

\begin{equation}
    f(x) = \frac{x}{1 + e^{-2\alpha x + \beta}}
\end{equation}

Then, suppose a large value of $\alpha$ and Equation (A5) is expressed as the follows:

\begin{equation}
     f(x) = \lim_{\alpha \rightarrow \infty} \frac{x}{1 + e^{-2\alpha x + \beta}}
\end{equation}

To clarify, we consider two cases of (1) $x \geq 0$, and (2) $x < 0$. In the first case, $e^{-2\alpha x + \beta}$ is converged to 0, and thus $f(x) = x$. In contrast, in the second case, $e^{-2\alpha x + \beta}$ is diverged, and thus $f(x) = 0$. Therefore, $f(x)$ is approximated as $\textit{max}(0, x)$. Here, since the definition of ASH activation function is originally as below, ASH could be approximated as the following:

\begin{equation}
\begin{aligned}
  \textit{\AE}(x^{(i)}) &=
  \begin{cases}
  x^{(i)} & \text{if}\ x^{(i)} - \mu_X - z_k\sigma_X \geq 0, \\
  0 & \text{otherwise}
  \end{cases} \\
\\
    &= \textit{max}(0, x^{(i)} - \mu_X - z_k\sigma_X) \\
\\
    &= \frac{x^{(i)}}{1 + e^{-2\alpha(x^{(i)} - \mu_X - z_k\sigma_X)}}
\end{aligned}
\end{equation}

Supplementary Fig. \ref{fig:ash_versions} shows the various versions of ASH alongside the various values of $\alpha$. As the value of $\alpha$ increasing, ASH activation function ($xH(x-a)$) could be reasonable approximated as hyperbolic function ($0.5x(1+\tanh(\alpha(x-a))$).

\section*{Appendix H. Comparison of ASHs}
\begin{figure}[h!]
    \centering
    \includegraphics[width=1.00\linewidth]{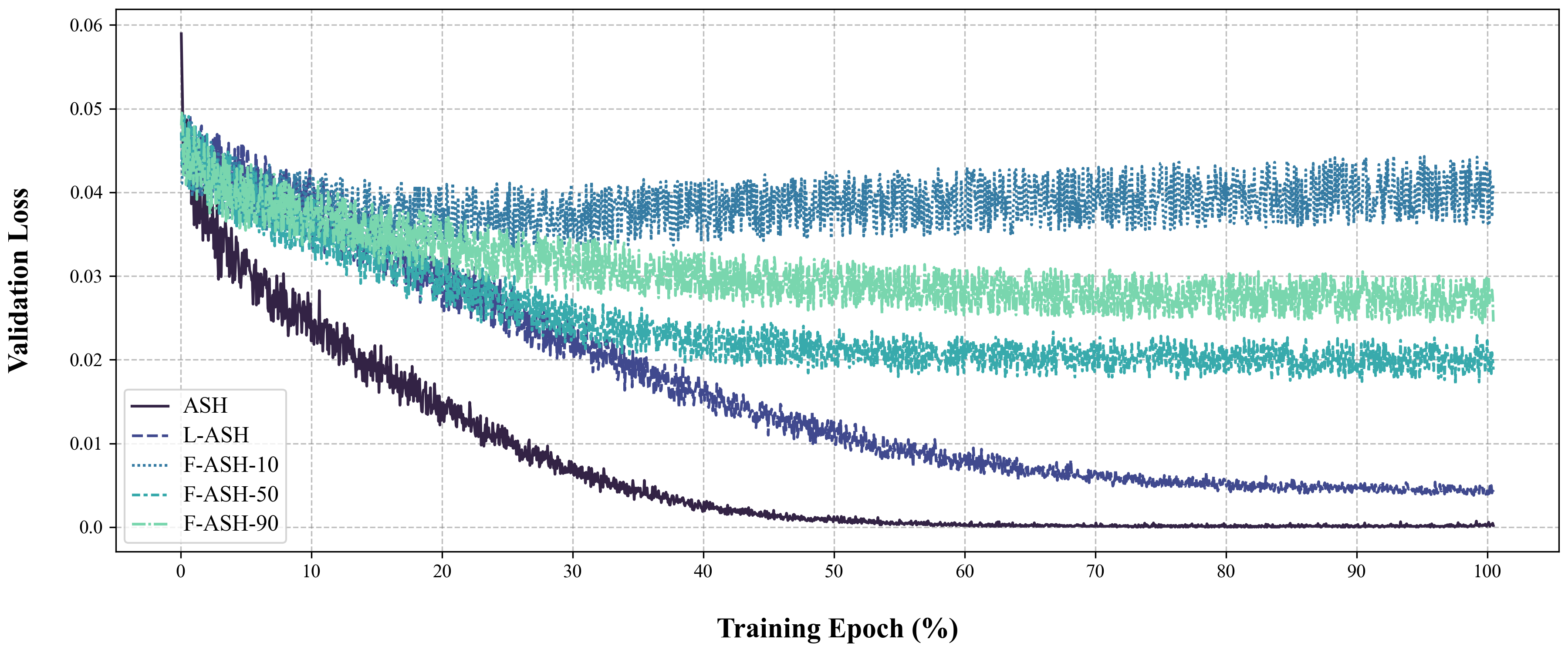}
    \caption{Comparison of various versions of ASH activation function. ASH activation function indicates the proposed activation function, L-ASH indicates Leaky ASH, and F-ASH-$k$ is ASH activation function that $k$-precentile is fixed rather than trainable.}
\label{fig:comp_ash}
\end{figure}

ASH activation function that rectified top-$k$\% percentile could be modified into various versions. Suppose Leaky ASH (L-ASH) that utilizes the scaling factor like Leaky ReLU, such that L-ASH is defined as follows:

\begin{equation}
  \AF(x^{(i)}) =
  \begin{cases}
  x^{(i)} & \text{if}\ x^{(i)}\ \text{is ranked in top}\ k\%\ \text{percentile of }\ X, \\
  ax & \text{otherwise}
  \end{cases}
\end{equation}

Here, $a$ is trainable like Leaky ReLU and positive value. L-ASH could improve the gradient vanishing problems. Additionally, we can employ fixed $k$\% percentile in ASH, and we denoted it as Fixed ASH with $k$ (F-ASH-$k$). F-ASH-$k$ rectifies the top-$k$ percentile from inputs but $k$ is fixed as a hyper-parameter. To find the best performance ASH, we compared the various version of ASH activation functions; ASH, L-ASH, F-ASH-10, F-ASH-50, and F-ASH-90. Supplementary Fig.~\ref{fig:comp_ash} shows the validation loss value alongside the training epoch for activation functions. The validation losses are averaged from the results of all experiments of classification, detection, and segmentation.

The experimental result illustrates that the proposed ASH activation function exhibits early convergence with lower validation loss values. Compared to ASH activation function, L-ASH exhibits slower convergence for the optimization due to the increased number of trainable parameters and decreased sparsity caused by $a$ in the negative domain below top-$k$\% percentile. Furthermore, F-ASH-$k$ activation functions exhibit lower performance in terms of validation loss since they rectify the fixed percentile of inputs regardless of contextual information of inputs. Here, F-ASH-50 shows slow but continuous convergence to the optimization since 50\% rectification could be regarded as ReLU activation function. In contrast, F-ASH-90 has limitations in the optimization since it could not localize the informative features in the inputs. Similarly, F-ASH-10 has limitations in the optimization since it extremely rectifies the overall information of inputs. The results conclude the benefits of ASH activation functions in terms of sparsity and trainable property.

\section*{Appendix I. Discussion}

\begin{figure}[h!]
    \centering
    \includegraphics[width=0.65\linewidth]{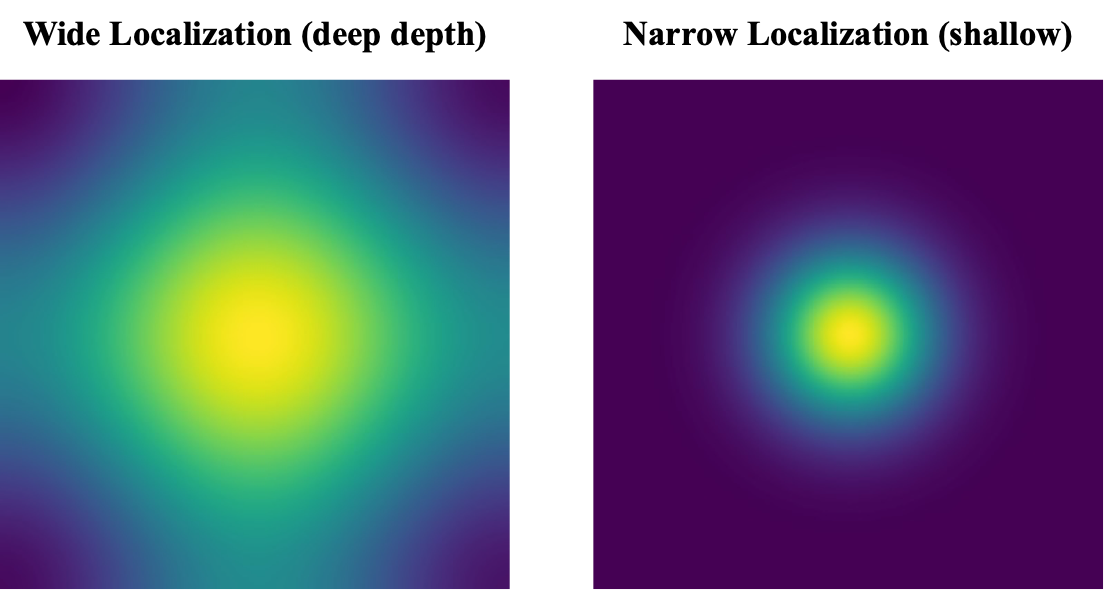}
    \caption{Localization property of ASH. Localization and pass out are realized in the wide fields in ASH activation function in deeper depth. Narrow localization and extreme rectification are realized in ASH activation function in shallow depth. }
\label{fig:localization_ash}
\end{figure}

\textbf{Localization property of ASH} $\qquad$ Supplementary Fig. \ref{fig:localization_ash} illustrates the localization example of ASH activation function. $z_k$ of ASH in a deeper depth exhibited smaller values (even negative values) compared to $z_k’$ of another ASH in a shallower depth. Note that, a small value of $z_k$ implies the wide range of top-$k$\% percentile, and thus it leads to the wide field of localization. In contrast, a large value of $z_k$ implies the narrow range of top-$k$\% percentile, and thus it provides extreme rectification and a narrow field of localization. Again, in the early layer, ASH activation function provides sufficiently broad but informative activations to forward layers. Additionally, ASH at the end of the models exhibits the activations that focus on the target object. Therefore, ASH activation function could provide informative activations to the deep learning models. Note that, the deep depth layers indicate that the layers are close to the input layer, whereas the shallow depth layers indicate that the layers are close to the output layers.

\textbf{Parameter Selection of ASH} $\qquad$ 
In this study, we proposed ASH activation function of which parameter $z_k$ is trainable and regarded as the hyper-parameter. However, ASHs exhibit significantly different $z_k$ values with significant variations even in the same networks. In addition, different values of $z_k$ are utilized concerning different tasks. Intuitively, a wide range of activations is required to recognize entire contexts of an image in a deeper depth, whereas a narrow range of activations is required to rectify features in a shallower depth in image recognition, especially a classification task. In contrast, consistent activation of the target object is required, in the segmentation or detection tasks. For instance, ASH in the initial position of the network exploits small $z_k$ (e.g., 90\% percentile sampling), whereas ASH at the end of the network exhibits an enormous $z_k$ value (e.g., 15\% percentile sampling). Therefore, rather than searching for the best performing parameters, we assumed that learning naturally from the network itself imitates more like human neurons. Thus, we skipped exploring the best parameter for ASH, in terms of $z_k$.

\textbf{Future Work} $\qquad$ Despite the experimental results demonstrating the superior performance of ASH activation function in accuracy, sparsity, training time, and localization property, the mathematical proofs of those properties are limited. The supporting mathematical analysis and proofs could be more discussed as potential future work. Furthermore, we mainly applied ASH activation functions to the task in which the role of the activation function is significantly issued, including classification, detection, segmentation, and image generation task. Since vision-based analysis could exhibit the properties of activation functions in a visual manner alongside the mathematical proofs, we mainly utilized the vision-based tasks. Text- and transformer-based analysis could be further discussed for future work. Furthermore, ASH activation function could be applied in the field of Natural Language Processing or Signal Compression field rather than in vision-based applications. This also remains as future work.


\end{document}